\documentclass{article}

\PassOptionsToPackage{compress, numbers}{natbib}
\usepackage[main, final]{neurips_2026}

\usepackage[utf8]{inputenc} 
\usepackage[T1]{fontenc}    
\usepackage{url}            
\usepackage{booktabs}       
\usepackage{amsfonts}       
\usepackage{nicefrac}       
\usepackage{microtype}      
\usepackage{amsmath}
\usepackage{amssymb}
\usepackage{amsthm}
\usepackage{graphicx}
\usepackage{float}
\newtheorem{assumption}{Assumption}
\newtheorem{theorem}{Theorem}
\newtheorem{proposition}{Proposition}
\newtheorem{lemma}{Lemma}
\newtheorem{corollary}{Corollary}

\usepackage[table,usenames,dvipsnames]{xcolor}
\usepackage[colorlinks,linktoc=all]{hyperref}
\hypersetup{citecolor=MidnightBlue}
\hypersetup{linkcolor=MidnightBlue}
\hypersetup{urlcolor=MidnightBlue}

\newtheorem{reptheorem}{Theorem}

\newcommand{\Comp}[1]{\mathrm{Comp}(#1)}
\title{Parallel Broyden Methods for Efficiently Evaluating Nonlinear State Space Models}

\author{%
  Ian Christopher Tanoh \\
  Department of Statistics\\
  Stanford University\\
  \texttt{ictanoh@stanford.edu} \\
  \And
  Scott W. Linderman \\
  Department of Statistics \\
  Stanford University \\
  \texttt{scott.linderman@stanford.edu} \\
}

\begin{document}

\maketitle

\begin{abstract}

Nonlinear state space models are typically evaluated sequentially, limiting their ability to exploit modern parallel hardware. Recent work has shown that state space models can be evaluated in parallel by reformulating evaluation as a root-finding problem, which can be solved with a parallel form of Newton's method. However, exact Newton methods require expensive multiplication of dense Jacobian matrices, and quasi-Newton methods based on diagonal approximations, while more scalable, are often slow to converge because they neglect interactions across state dimensions. We introduce a novel quasi-Newton approach based on Broyden's method, which captures coupling terms with diagonal-plus-low-rank approximations to blocks of the Jacobian. We construct the approximations from trajectory secants, avoiding automatic differentiation while retaining the favorable scaling of diagonal approximations and remaining compatible with parallel scan through rank compression. We prove finite-step recovery of the exact trajectory under the compressed scan, and show that convergence is tracked by the compression error and by the block approximation error along the current trajectory error, rather than in spectral norm. Empirically, we show that our method is advantageous when interactions are approximately low rank or when Jacobian evaluation is expensive: on low-rank RNNs, parallel MCMC algorithms, and denoising diffusion models, it converges in
fewer iterations than other quasi-Newton methods and is the fastest parallel
method when Jacobian evaluation is expensive or state dimensions are large.
\end{abstract}

\section{Introduction}

Parallel computation has been a key driver of recent advances in machine learning. Architectures such as transformers \citep{vaswani2017attention} and linear state space models \citep{gu2021efficiently, gu2023modeling,  smith2022simplified, martin2017parallelizing} are explicitly designed to exploit parallelism over sequence length. In contrast, general nonlinear state space models, including recurrent neural networks \citep{graves2012long,cho2014learning,  mienye2024recurrent}, diffusion samplers \citep{song2020denoising, song2020score, karras2022elucidating, ho2020denoising}, and Markov chain Monte Carlo methods \citep{geman1984stochastic, hastings1970monte, metropolis1953equation, neal2011mcmc, roberts1998optimal}, seem inherently sequential, requiring time that scales linearly with trajectory length. This sequential dependency limits their ability to exploit modern parallel hardware.

Recent work, including DEER~\citep{lim2023parallelizing} and DeepPCR~\citep{danieli2023deeppcr}, has addressed this challenge by reformulating trajectory computation as a root-finding problem over the full sequence. In this formulation, the residual Jacobian has a block-bidiagonal structure, so Newton updates reduce to solving a linear recurrence that can be evaluated with logarithmic parallel depth via parallel scan~\citep{blelloch1989scans}. This makes it possible to replace sequential evaluation with iterative parallel trajectory refinement.

However, exact Newton updates face two scalability bottlenecks. First, they require storing and multiplying dense dynamics Jacobian matrices, leading to memory and computational costs that scale poorly with state dimension. Second, evaluating these Jacobians with automatic differentiation can be expensive when the transition map is a large neural network or simulator. Scalable quasi-Newton variants address the first bottleneck by replacing each block with a diagonal approximation, reducing memory and computation to scale linearly with state dimension \citep{gonzalez2024towards}. However, these methods still require estimating or computing the diagonal of the Jacobian, and therefore do not fully remove the cost of Jacobian evaluation. Moreover, the diagonal approximation discards dependencies across state dimensions, which can hinder convergence when the dynamics exhibit strong coupling \citep{gonzalez2025unifying}.

In this work, we ask: \emph{Can we construct richer Jacobian approximations that retain the scalability of diagonal quasi-Newton while capturing structured cross-dimensional coupling and avoiding expensive Jacobian evaluation?} We answer this question by introducing Parallel Broyden methods, a family of quasi-Newton methods based on Broyden's method that can be parallelized over the sequence length. Broyden's method is a classical quasi-Newton approach for updating Jacobian approximations from secant information~\citep{broyden1965class}, without automatic differentiation.  Broyden-style updates underlie several widely used optimization methods, including BFGS and L-BFGS~\citep{broyden1970convergence,nocedal1980updating}. Our approach targets regimes where cross-dimensional interactions are important but can be captured by low-rank corrections, or where automatic differentiation through the transition map is costly.

Our contributions are as follows:
\begin{itemize}
    \item We introduce Parallel Broyden, a family of quasi-Newton methods for parallel nonlinear trajectory evaluation that build scan-compatible diagonal-plus-low-rank Jacobian approximations from trajectory secants, avoiding costly automatic differentiation.

    \item We develop a compressed approximate scan that controls the rank of diagonal-plus-low-rank blocks while preserving logarithmic parallel depth and scaling linearly with state dimension for small rank. We prove finite-step recovery of the exact trajectory even with compression.

    \item We extend the convergence analysis of parallel quasi-Newton methods to
compressed scans, and show theoretically and empirically that convergence is
tracked by the block approximation error along the current trajectory error, rather than
in spectral norm.

    \item We show empirically on low-rank RNNs and hierarchical Gibbs sampling that
our approach captures low-rank coupling more effectively than diagonal
quasi-Newton, and on diffusion models that it avoids costly differentiation
through the denoiser, reducing iteration counts relative to diagonal
quasi-Newton and improving runtime over exact and diagonal methods when
Jacobian computation or scaling costs dominate, with speedups of up to
$30\times$ over sequential evaluation. Code is available at \url{https://github.com/ianctanoh/BroyDEER}.
\end{itemize}

\section{Problem Statement and Related Work}

Parallel computation over sequences is central to modern sequence modeling.
Structured state space models such as S4, S5, and Mamba exploit algebraic
structure for efficient long-sequence computation
\citep{gu2021efficiently,smith2022simplified,gu2023modeling}. For general
nonlinear recurrences, recent work instead recasts trajectory evaluation as a
nonlinear system solved by iterative refinement, including DEER and
DeepPCR~\citep{lim2023parallelizing,danieli2023deeppcr}, Picard iteration for parallel diffusion sampling ~\citep{shih2023parallel,tang2024accelerating}, parallel-in-time
methods for ODE solvers \citep{iacob2025parallel}, and MCMC
\citep{zoltowskiparallelizing}. Recent unifying work views
these approaches as quasi-Newton iterations on the full trajectory residual
\citep{gonzalez2025unifying}. We follow this perspective, using Broyden-style
secant updates to obtain scalable quasi-Newton approximations for parallel
trajectory evaluation.

Following the notation of \citet{gonzalez2025unifying}, we consider a
$D$-dimensional nonlinear state-space model with time-indexed transition maps
$f_t : \mathbb{R}^D \to \mathbb{R}^D$, where the state at time $t$ is
$x_t \in \mathbb{R}^D$. The maps $f_t$ may represent, for example, the transition
of a recurrent neural network, one step of a numerical simulator, a sampling
update, or a diffusion model denoising step. Given an initial state $x_0$, our
goal is to compute the trajectory $\mathbf{x}_{1:T} \in \mathbb{R}^{T \times D}$
satisfying
\begin{equation}
    x_t = f_t(x_{t-1}), \quad t = 1, \ldots, T.
    \label{eq:dynamics}
\end{equation}
Although this recurrence is naturally evaluated sequentially, our aim is to
expose structure that permits parallel computation across the time axis. For
clarity, we absorb input dependence into the time-indexed transition maps
$f_t$.

\paragraph{Residual formulation.}
Rather than solving \eqref{eq:dynamics} sequentially, we recast trajectory
computation as a root-finding problem. For any candidate trajectory $\mathbf{x}_{1:T}$,
define the block residual
\begin{equation}
    r_t(\mathbf{x}_{1:T}) := x_t - f_t(x_{t-1}), \quad t = 1, \ldots, T,
    \label{eq:block_residual}
\end{equation}
and stack these blocks into $\mathbf{r}(\mathbf{x}_{1:T}) \in \mathbb{R}^{T \times D}$. The true
trajectory $\mathbf{x}^\star_{1:T}$ is characterized by
\begin{equation}
    \mathbf{r}(\mathbf{x}^\star_{1:T}) = 0,
    \label{eq:root}
\end{equation}
with each residual block measuring the one-step prediction error at time $t$.
Our goal is to solve \eqref{eq:root} in sublinear parallel depth while keeping
the per-iteration cost favorable in the state dimension.

\paragraph{Exact parallel Newton \citep{lim2023parallelizing, danieli2023deeppcr}.}
The root-finding formulation \eqref{eq:root} allows trajectory evaluation to be
performed by Newton iteration over the full sequence. Starting from a candidate
sequence $\mathbf{x}^{(i)}_{1:T}$, Newton's method computes a correction
$\Delta \mathbf{x}^{(i)}$ that locally reduces all residuals
$r_1,\ldots,r_T$ simultaneously by solving
\begin{equation}
    \mathbf{J}(\mathbf{x}^{(i)}_{1:T}) \, \Delta \mathbf{x}^{(i)} = -\mathbf{r}(\mathbf{x}^{(i)}_{1:T}),
    \label{eq:lds}
\end{equation}
where $\mathbf{J}(\mathbf{x}^{(i)}_{1:T})$ is the Jacobian of the residual map with respect to the
trajectory variables. This Jacobian has special structure because each residual
block $r_t = x_t - f_t(x_{t-1})$ depends only on the current and previous states.
Treating $\mathbf{x}_{1:T}$ as a flattened vector in $\mathbb{R}^{TD}$ and defining
\[
    A_t^{(i)}
    :=
    \frac{\partial f_t}{\partial x_{t-1}}
    \left(x^{(i)}_{t-1}\right)
    \in \mathbb{R}^{D \times D},
\]
the residual Jacobian is block lower bidiagonal:
\[
    \mathbf{J}(\mathbf{x}^{(i)}_{1:T}) := \frac{\partial \mathbf{r}}{\partial \mathbf{x}_{1:T}}(\mathbf{x}^{(i)}_{1:T}) =
    \begin{pmatrix}
        I_D      & 0        & \cdots & 0          & 0      \\
        -A_2^{(i)}     & I_D      & \cdots & 0          & 0      \\
        0        & -A_3^{(i)}     & \ddots & 0          & 0      \\
        \vdots   & \vdots   & \ddots & I_D        & 0      \\
        0        & 0        & \cdots & -A_T^{(i)}       & I_D
    \end{pmatrix}.
\]
Solving \eqref{eq:lds} is therefore equivalent to evaluating a linear recurrence over
time, which can be computed with logarithmic parallel depth using parallel scan, a standard primitive for prefix computation.

\paragraph{Diagonal quasi-Newton \citep{gonzalez2024towards}.}
Exact Newton is costly in high dimensions because it must form, store, and
compose dense dynamics Jacobian blocks $A_t^{(i)}$ inside the parallel scan,
leading to $\mathcal{O}(TD^2)$ memory and $\mathcal{O}(TD^3)$ work. Diagonal quasi-Newton
reduces this cost by replacing each block with a diagonal approximation. Since
diagonal matrices are closed under multiplication, the scan remains exact under
the approximation while reducing memory and computation to scale linearly in
$D$ per block. In practice, this approach is most effective when the dynamics
Jacobian is well approximated by its diagonal and when the cost of dense
Newton-style updates is prohibitive at scale \citep{gonzalez2025unifying}.

\paragraph{Beyond the diagonal approximation.}
However, the diagonal approximation may become too crude when the dynamics are
strongly coupled across dimensions. This occurs, for example, in hierarchical
Gibbs samplers, where updating one latent variable can change the conditional
distribution of many others. Moreover, \citet{gonzalez2025unifying} bound the convergence rate of quasi-Newton updates in terms of the spectral-norm error of the block Jacobian approximation, suggesting that more accurate block approximations lead to
faster convergence. This motivates the search for richer
quasi-Newton block approximations that better capture these interactions while
retaining the $\mathcal{O}(TD)$ memory and work of diagonal quasi-Newton methods.

\section{Parallel Broyden methods for scalable sequence evaluation}

We now introduce Broyden-based quasi-Newton approximations for parallel nonlinear trajectory evaluation. Figure~\ref{fig:overview} (Appendix~\ref{app:overview}) gives an overview of the method. The methods, which we refer to as Parallel Broyden methods, construct diagonal-plus-low-rank (DPLR) approximations of local dynamics Jacobian blocks from trajectory secants rather than automatic differentiation. We first describe the rank-one Broyden updates underlying this construction, then we present a multi-secant variant that we call Accelerated Parallel Broyden. We then show how these DPLR blocks can be made compatible with logarithmic-depth parallel scan through rank compression. Finally, we show that the compressed scan preserves finite-step recovery of
the exact trajectory, and analyze how block approximation error and rank
compression affect the local convergence rate.

\subsection{DPLR Jacobian approximation with Broyden's method}
\label{sec:broyden}

Parallel Broyden applies Broyden's method blockwise to the local
dynamics functions. Rather than forming Jacobians with automatic differentiation, it
updates each block approximation using finite differences between successive
trajectory iterates. Since Broyden updates are rank-one corrections, initializing
with a diagonal baseline naturally yields diagonal-plus-low-rank approximations
without automatic differentiation.

\paragraph{Blockwise Broyden update.}
After computing the iterate $\mathbf{x}^{(i+1)}_{1:T}$, we form a secant
pair for each transition $f_t$:
\[
    p_t^{(i)} := x_{t-1}^{(i+1)} - x_{t-1}^{(i)},
    \qquad
    q_t^{(i)} :=
    f_t(x_{t-1}^{(i+1)}) - f_t(x_{t-1}^{(i)}).
\]
The new approximation $\widetilde A_t^{(i+1)}$ is then required to satisfy the \textit{secant condition}:
$$\widetilde A_t^{(i+1)} p_t^{(i)} = q_t^{(i)}.$$
This is the multidimensional analogue of a finite-difference estimate of a
derivative. Since the condition does not uniquely determine
$\widetilde A_t^{(i+1)}$, Broyden's method chooses the feasible matrix closest to
the current approximate dynamics Jacobian $\widetilde A_t^{(i)}$:
\begin{equation*}
    \widetilde A_t^{(i+1)}
    =
    \arg\min_{A} \|A - \widetilde A_t^{(i)}\|_F
    \quad \text{s.t.} \quad
    A p_t^{(i)} = q_t^{(i)}.
\end{equation*}
The solution is a rank-one correction to the previous approximation (see Appendix \ref{app:broyden_standard}).

\paragraph{DPLR structure and limited memory.}
Each Broyden update adds a rank-one correction to the current approximation.
Starting from a diagonal initialization
$\widetilde A_t^{(0)} = D_t^{(0)}$,
after $i$ updates the approximation can be written as
\begin{equation*}
    \widetilde A_t^{(i)}
    =
    D_t^{(0)} + U_t^{(i)} {V_t^{(i)}}^\top,
    \qquad
    U_t^{(i)}, V_t^{(i)} \in \mathbb{R}^{D \times i}.
\end{equation*}
Each column pair in $U_t^{(i)}$ and $V_t^{(i)}$ corresponds to one secant
correction. In our experiments, we choose
$D_t^{(0)}=\gamma I_D$, with $\gamma$ a task-dependent hyperparameter.
Since the rank of the low-rank component grows with the number of Broyden
updates, we use a limited-memory variant that retains only the most recent $m$
secant corrections per block. This yields a DPLR approximation with rank at most $m$ while keeping the representation bounded and compatible with the rank-truncated scan described below. We also considered an L-BFGS-style variant that rescales the diagonal baseline from the most recent secant pair before replaying the stored corrections. Since this added complexity did not substantially change performance in our experiments, we use the simpler fixed-baseline truncation as the default (see Appendix \ref{app:broyden_limited_memory}). 

\begin{table}[t]
\centering
\caption{Comparison of trajectory evaluation methods. Sequential is the baseline. All other methods are parallel via parallel scan with $\mathcal{O}(\log T)$ depth. Parallel Broyden and Accelerated Parallel Broyden (this work) use diagonal-plus-low-rank (DPLR) block approximations with rank at most $m$.}
\label{tab:methods}
\resizebox{\textwidth}{!}{%
\begin{tabular}{lcccc}
\toprule
Method & Block approximation $\widetilde A_t^{(i)}$ & Work & Memory & Uses Autodiff \\
\midrule
Sequential  & --- & $\mathcal{O}(TD^2)$  & $\mathcal{O}(D)$  & No\\

Newton \cite{lim2023parallelizing, danieli2023deeppcr}
& $A_t^{(i)}=\dfrac{\partial f_t}{\partial x_{t-1}}(x_{t-1}^{(i)})$
& $\mathcal{O}(TD^3)$
& $\mathcal{O}(TD^2)$
& Yes\\

Diagonal quasi-Newton \cite{gonzalez2024towards}
& $\operatorname{diag}\!\left(A_t^{(i)}\right)$
& $\mathcal{O}(TD)$
& $\mathcal{O}(TD)$
& Yes\\

Parallel Broyden (this paper)
& $D_t^{(0)} + U_t^{(i)}{V_t^{(i)}}^\top$, $\operatorname{rank}\le m$
& $\mathcal{O}(TDm^2)$
& $\mathcal{O}(TDm)$
& No \\

Accelerated Parallel Broyden (this paper)
& $D_t^{(0)} + U_t^{(i)}{V_t^{(i)}}^\top$, $\operatorname{rank}\le m$
& $\mathcal{O}(TDm^2 + Tm^3)$
& $\mathcal{O}(TDm)$
& No \\

\bottomrule
\end{tabular}%
}
\end{table}

\paragraph{Multi-secant variant.}
The rank-one Broyden update incorporates one secant condition at a time. As a standard extension, we also consider a multi-secant variant that uses the most recent $m$ trajectory secants simultaneously \citep{fang2009two,schnabel1983quasi}. Rather than accumulating rank-one corrections over time, this variant fits a rank-at-most-$m$ correction in a single step on top of a reference matrix. In our experiments, we take this reference matrix to be $\gamma I_D$, resulting in a DPLR approximation with correction at most rank $m$. We call this the Accelerated Parallel Broyden method. It adds only small $m \times m$ linear solves and, as we will see, can substantially reduce the number of iterations in practice. Details are deferred to Appendix~\ref{app:broyden_multisecant}. When needed for stability, we damp the low-rank corrections (Appendix~\ref{app:broyden_damping}). The values used in each experiment are
given in the corresponding appendices.

\subsection{Scan-compatible DPLR approximations}
\label{sec:dplr}

The Broyden updates above naturally produce DPLR
approximations of the local dynamics Jacobian blocks. Specifically, we represent
each approximate block as $\tilde{A}_t = \gamma I_D + U_t V_t^\top$ with $U_t, V_t \in \mathbb{R}^{D \times m}$, and $m \ll D$. The main challenge is compatibility with the parallel scan. Diagonal matrices
are closed under multiplication, so their representation remains fixed across
the scan tree. In contrast, the product of two DPLR matrices gives a diagonal matrix plus a correction of rank at most
$2m$, so repeated composition can eventually lead to a full rank matrix.

We address this by introducing a novel \emph{compressed approximate scan}, which allows
DPLR matrices to remain compatible with parallel
scan by controlling rank growth at each merge. Instead of requiring closure
under composition, we allow rank growth at each merge and compress the resulting
correction of rank at most $2m$ back to rank at most $m$. This keeps the
representation size bounded while preserving $\mathcal{O}(\log T)$ parallel
depth. Because compression is not exact, the merge is only approximately associative,
so the result can depend on the order in which products are formed and
compressed, i.e., on the scan tree. At the ranks used in our experiments, the
choice of tree has no meaningful effect on convergence
(Appendix~\ref{app:scan_tree}), and all experiments use the Blelloch-type scan
implemented by \texttt{jax.lax.associative\_scan}. Note that in the purely low-rank case with no diagonal term, the class of rank-at-most $m$ matrices is closed under multiplication, so no compression is needed.

\paragraph{Randomized rank compression.}
We compress directly in factorized form using randomized low-rank approximation
\citep{saha2023matrix,witten2015randomized}. Given a rank-at-most-$2m$
correction $UV^\top$ with $U,V\in\mathbb{R}^{D\times 2m}$, draw
$\Omega\in\mathbb{R}^{D\times m}$ and form
$Y=(UV^\top)\Omega=U(V^\top\Omega)$. Let $Y=QR$ be a thin QR decomposition.
Projecting onto $\mathrm{span}(Q)$ gives
$  UV^\top \approx Q(Q^\top U)V^\top
    = \widetilde U\widetilde V^\top$,
with 
    $\widetilde U=Q$ and $\widetilde V=V(U^\top Q)$.
This avoids dense matrices and singular value decomposition (SVD).
As a reference, we also implement exact compression via a small $2m\times 2m$ SVD. Appendix~\ref{app:compression} shows that rank compression has little effect on convergence rate at our ranks, while the randomized scheme is fastest.


\paragraph{Complexity.}
For fixed rank $m$, storing a DPLR block requires $\mathcal{O}(Dm)$ memory and
applying it costs $\mathcal{O}(Dm)$. Composition followed by randomized
compression costs $\mathcal{O}(Dm^2)$. Across all $T$ blocks, this yields
$\mathcal{O}(TDm)$ memory while preserving the $\mathcal{O}(\log T)$ parallel
depth of the scan. Thus, small $m$ preserves linear scaling in $D$ while extending beyond diagonal quasi-Newton (Table \ref{tab:methods}).

\paragraph{Global convergence.}
A key question is whether the compressed approximate scan preserves the finite-step recovery property of parallel (exact or quasi) Newton methods: after $i$ iterations, the first $i$ states are recovered exactly, and the full trajectory is recovered in at most $T$ iterations~\citep{gonzalez2024towards}. The answer is yes. Although compression changes the matrix factors used inside the scan, the inductive recovery argument remains valid because correct prefixes have zero residual offsets.

\begin{theorem}[Global convergence with approximate scan]
\label{thm:global_recovery}
Starting from any initial trajectory $\mathbf{x}^{(0)}_{1:T}$, after $i$
iterations using the approximate scan, we have $x_t^{(i)} = x_t^\star$ for all
$t \leq i$. In particular, the method recovers the exact trajectory
$\mathbf{x}^\star_{1:T}$ in at most $T$ iterations (Proof in Appendix \ref{app:global_recovery}).
\end{theorem}

\subsection{Effect of Jacobian approximation and compression error on convergence}
\label{sec:convergence}

Theorem~\ref{thm:global_recovery} guarantees recovery in at most $T$
iterations, but says nothing about how quickly the error decreases. For
quasi-Newton updates with an \emph{exact} scan,
\citet[Prop.~3]{gonzalez2025unifying} show that, to first order, each iteration
contracts the error by a factor
$\|\widetilde{\mathbf J}^{-1}\|_2\,\max_t\|\widetilde A_t-A_t\|_2$. We extend
this result to obtain rates independent of $T$ and to account for rank
compression inside the scan.

\paragraph{Assumptions.}
Following \citet{gonzalez2025predictability}, convergence is governed by
products of Jacobians: if they contract, errors remain local and parallel Newton
converges rapidly. If they grow, errors propagate over long horizons. Since the
solver multiplies its approximate blocks rather than the true Jacobians, we
assume products $\widetilde A_{t+k}\cdots\widetilde A_{t+1}$ grow at most at an
exponential rate $\tilde\lambda$, up to a transient factor $\tilde a\geq1$. This
rate is effectively the largest Lyapunov exponent (LLE) of the approximate
system. We further assume $L$-Lipschitz Jacobians, that the blocks are within $\rho$ of the true Jacobians along the current trajectory error, and that each compression step in the scan introduces a relative error of at most $\eta$ (Appendix~\ref{app:conv}).

\begin{theorem}[Local convergence with compressed scan]
\label{thm:cvg_bound}
Suppose the assumptions above hold at every iteration $i\geq i_0$. Then:
\begin{enumerate}
    \item[(i)] \emph{Effective predictability.} Compression raises the
    effective LLE from $\tilde\lambda$ to
    $\lambda_c:=\tilde\lambda+\log(1+\tilde a^2\eta)$.

    \item[(ii)] \emph{One-step contraction.} Suppose $\lambda_c<0$, and write
    $\|e\|_\infty:=\max_t\|e_t\|_2$. Then for every iteration $i\geq i_0$,
    \[
        \|e^{(i+1)}\|_\infty\le\big(c_0+c_1\|e^{(i)}\|_\infty\big)\|e^{(i)}\|_\infty,
    \]
    where $c_0$ and $c_1$ are independent of $T$, with
    \[
        c_0:=\frac{\tilde a\,\rho}{1-e^{\lambda_c}}
        +\frac{2\tilde a^3\eta\,e^{\tilde\lambda}}{(1-e^{\tilde\lambda})(1-e^{\lambda_c})},
        \qquad
        c_1:=\frac{\tilde a\,L}{2(1-e^{\lambda_c})}.
    \]

    \item[(iii)] \emph{Local linear convergence.} If $c_0<1$ and
    $\|e^{(i_0)}\|_\infty<(1-c_0)/c_1$, then
    $\|e^{(i)}\|_\infty\le q^{\,i-i_0}\|e^{(i_0)}\|_\infty$ for all $i\geq i_0$, with
    $q:=c_0+c_1\|e^{(i_0)}\|_\infty<1$ (proof in Appendix~\ref{app:conv}).
\end{enumerate}
\end{theorem}

\paragraph{Implications.}
First, for exact parallel Newton ($\rho=\eta=0$), $c_0=0$ and (ii) gives quadratic convergence. Any approximation makes $c_0>0$, so our bound guarantees only a linear rate,
even with exact Jacobians under compression. Second, the result is local: once
the error falls below $(1-c_0)/c_1$, the iteration count is independent of $T$,
giving $O(\log T)$ depth. We do not bound how long entering this region takes.
When $\lambda_c\geq0$, it may not exist or may shrink with $T$. Third, whereas
prior analyses measure the quality of the block approximation by its
spectral-norm error \citep{gonzalez2025unifying}, our bound depends on the block error along the current trajectory error, $\rho$,
which is never larger than the spectral-norm error. Experiments on low-rank RNNs and hierarchical Gibbs
sampling suggest that $\rho$ is the quantity that matters
(Appendix~\ref{app:conv_diagnostics}): the spectral-norm error is essentially the same for all three methods, whereas near convergence $\rho$ is ordered as Accelerated Parallel Broyden $<$ Parallel Broyden $<$ diagonal quasi-Newton, matching their iteration counts, and damping helps Parallel Broyden by reducing $\rho$ but not the spectral-norm error. We note that the bound in (ii), however, is not tight.

\section{Experiments}

We study the efficacy of Parallel Broyden methods in three examples and find that for large enough state sizes and long enough sequences, these methods can be faster than sequential and parallel (quasi-) Newton methods.

\subsection{Controlled experiments on RNNs}

We begin with controlled experiments designed to understand when and why Parallel Broyden methods help, by varying the Jacobian's structure and the cost of computing it independently.
Our starting point is a family of low-rank RNNs whose Jacobians are exactly DPLR.
We consider dynamics of the form,
\[
    x_t
    =
    a x_{t-1}
    +
    \frac{\Delta t}{\tau} M \psi\!\left(N^\top \phi(x_{t-1})\right)
    +
    \epsilon_t,
\]
where $x_t \in \mathbb{R}^D$, $a = 1 - \Delta t/\tau$,
$\phi(\cdot)=\tanh(\cdot)$ is applied elementwise,
$M,N \in \mathbb{R}^{D \times R}$ with $R \ll D$, and $\psi:\mathbb{R}^R \to \mathbb{R}^R$ acts in the $R$-dimensional bottleneck.
Fixing the noise sequence $\{\epsilon_t\}$ turns the model into a deterministic input-driven state-space model. We consider two choices of $\psi$: the standard low-rank RNN, widely used in computational neuroscience~\citep{pals2024inferring,schuessler2020interplay,mastrogiuseppe2018linking}, with $\psi(z) = z$, and a variant we construct for this study, the MLP-RNN, with $\psi(z) = z + \beta\,\mathrm{MLP}(z)$, where the MLP has $8$ hidden layers of width $D$ and $\beta = 0.5$. In both cases, the transition Jacobian
$A_t(x_{t-1}) = a I_D + \frac{\Delta t}{\tau} M\, J_\psi\!\left(N^\top \phi(x_{t-1})\right) N^\top \operatorname{diag}(\phi'(x_{t-1}))$
is diagonal plus rank at most $R$. The true Jacobian therefore lies exactly in
the approximation class targeted by Parallel Broyden, whereas diagonal quasi-Newton
discards the low-rank coupling term.
In these experiments, we treat the dynamics $f$ as a black box, as in DEER~\citep{lim2023parallelizing}: methods may evaluate $f$, and Newton and diagonal quasi-Newton obtain exact Jacobian information (the full Jacobian, or its exact diagonal) by automatic differentiation of $f$, rather than from closed-form expressions specific to the model or from approximations to it. This keeps the comparison general: closed-form Jacobians are rarely available for learned or complex dynamics, and approximate Jacobian information can be unreliable on some problems (cf. Section \ref{sec:gibbs}). Under this assumption, Jacobian information is cheap for the standard low-rank RNN but expensive for the MLP-RNN, where it requires differentiating through the MLP. Both Broyden variants, in contrast, only require evaluations of $f$.
We use $R=3$, $\tau=10$, $D=200$, and Broyden rank $m=R$. Details are given in
Appendix~\ref{app:low_rank_rnn}.

\begin{figure}
    \centering
    \includegraphics[width=1.\linewidth]{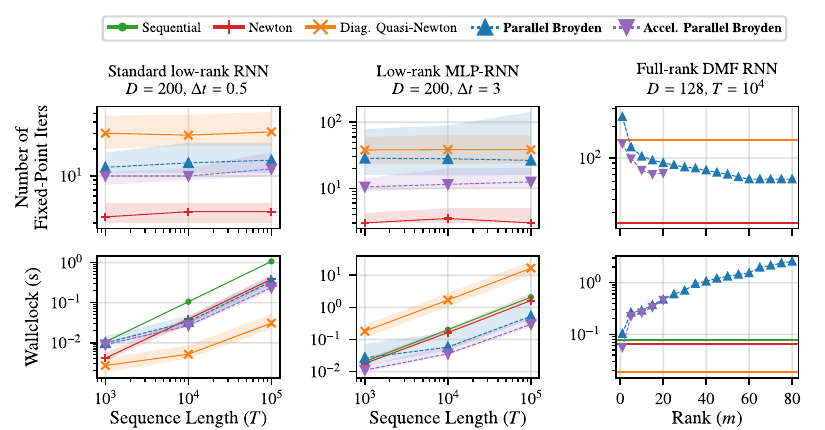}
        \caption{
\textbf{Low-rank and full-rank RNNs.}
\textbf{Left, middle:} Standard ($\Delta t = 0.5$) and MLP ($\Delta t = 3$) low-rank RNNs ($R=3$, $D=200$), as $T$ varies.
In both, iteration counts are ordered as diagonal quasi-Newton $>$ Parallel Broyden $>$ Accelerated Parallel Broyden $>$ Newton.
This ordering translates into runtime only when Jacobian information is expensive (MLP-RNN).
\textbf{Right:} Full-rank DMF RNN ($D=128$, $T=10^4$) as the Broyden rank $m$ varies. Methods without a rank parameter are shown as horizontal lines.
Accelerated Parallel Broyden is shown only up to $m=20$, as for larger $m$ it becomes numerically unstable or fails to converge.
Larger $m$ reduces iterations but increases runtime.
We report median iteration counts and runtimes with interquartile ranges over $32$ random realizations. Runtimes are computed over realizations that converged within the $1000$ iterations budget.
}

    \label{fig:fig1}
\end{figure}

We assess convergence and runtime on $32$ random realizations as we vary the trajectory length $T$ (Figure~\ref{fig:fig1}, left and middle).
In both the standard and MLP models, iteration counts follow the same ordering at every $T$:
diagonal quasi-Newton $>$ Parallel Broyden $>$ Accelerated Parallel Broyden $>$ Newton.
Both Broyden variants thus capture much of the off-diagonal coupling that diagonal quasi-Newton misses, and the multisecant updates of Accelerated Parallel Broyden capture it even more effectively.
The gap between the two Broyden variants depends on the strength of the off-diagonal coupling relative to the diagonal.
With $\Delta t = 0.5$ ($a = 0.95$), Parallel Broyden needs only $1.25$--$1.4\times$ as many iterations as the accelerated variant.
With $\Delta t = 3$ ($a = 0.7$), it needs $2.1$--$2.7\times$ as many, and it more often fails to converge within $1000$ iterations ($5$ of $32$ realizations vs.\ at most $1$ of $32$ for the accelerated variant). The same pattern holds for each model at both values of $\Delta t$ (Appendix~\ref{app:low_rank_rnn}). Parallel Broyden therefore closes most of the gap with the accelerated variant when off-diagonal coupling is weak.

Whether this ordering carries over to runtime depends on the cost of Jacobian information.
For the standard low-rank RNN, differentiating the dynamics is so cheap that diagonal quasi-Newton iterations take about $1$\,ms, compared with about $20$\,ms for Accelerated Parallel Broyden at $T=100\mathrm{K}$.
Diagonal quasi-Newton is therefore the fastest method despite requiring $2$--$3\times$ more iterations, running about $7.7\times$ faster than Accelerated Parallel Broyden at $T=100\mathrm{K}$. For the MLP-RNN, the per-iteration cost of both Broyden variants changes little, whereas diagonal quasi-Newton becomes about $400\times$ more expensive per iteration, and Newton, whose cost is already dominated by its dense $D \times D$ scan, becomes $5$--$6\times$ more expensive. The iteration ordering then translates directly into runtime, and Accelerated Parallel Broyden is the fastest method at every $T$.
At $T=100\mathrm{K}$, it is $7.0\times$ faster than sequential evaluation, $5.5\times$ faster than Newton, $1.8\times$ faster than Parallel Broyden, and $56\times$ faster than diagonal quasi-Newton.

Finally, we ask how our methods perform in an unfavorable case, where the Jacobian is far from DPLR. We consider a full-rank discretized mean-field (DMF) RNN~\citep{engelken2023lyapunov},
$x_t = W\tanh(x_{t-1}) + \epsilon_t$ with $W_{ij} \sim \mathcal{N}(0, g^2/D)$, $g=1.7$, and $D=128$. Its Jacobian is dense with no dominant diagonal, and its singular values decay slowly rather than sharply (Appendix~\ref{app:dmf_rnn}). We sweep the Broyden rank $m$ at $T=10\mathrm{K}$ (Figure~\ref{fig:fig1}, right). Performance degrades gracefully: iteration counts fall steadily with $m$ before plateauing, from $249$ at $m=1$ to $63$ for $m \geq 60$ for Parallel Broyden, and from $138$ to about $70$ by $m=15$ for the accelerated variant. The secant corrections therefore still capture real structure even when the spectrum does not decay sharply, and for $m \geq 5$ both variants need fewer iterations than diagonal quasi-Newton. Accelerated Parallel Broyden needs fewer iterations than Parallel Broyden at every rank where both are run, but it cannot use arbitrarily large $m$.
Beyond $m=20$, we observed that the most recent $m$ steps become nearly linearly dependent, so the Gram matrix of the multisecant update becomes ill-conditioned and the method becomes numerically very unstable. The question is whether the iterations saved by the secant corrections are worth their extra per-iteration cost. Because the per-iteration cost grows rapidly with $m$, the modest reduction in iterations does not pay off: runtime increases with $m$, and for $m \geq 5$ both variants are slower than every baseline. The fastest Broyden configuration (accelerated, $m=1$) is only $1.4\times$ faster than sequential evaluation, while diagonal quasi-Newton is fastest ($4.1\times$). Parallel Broyden variants are thus most advantageous when a small rank already improves substantially on diagonal quasi-Newton, which happens when the off-diagonal part of the Jacobian has sharply decaying singular values, as in the low-rank RNN.

\subsection{Gibbs sampling for hierarchical models}
\label{sec:gibbs}

Recently,~\citet{zoltowskiparallelizing} showed that parallel Newton methods can be applied to simulating Markov Chain Monte Carlo algorithms, leading to substantial improvements in runtime.
We show that Parallel Broyden methods can further accelerate performance.
We study the eight schools problem, a standard example in Bayesian hierarchical modeling~\citep{rubin1981estimation, gelman1995bayesian}. We observe test scores from $N$ students in each of $S$ schools according to the hierarchical model,
\[
    y_{s,n} \mid \theta_s,\sigma_s^2 \sim
    \mathcal{N}(\theta_s,\sigma_s^2),
    \qquad
    \theta_s \mid \mu,\tau^2 \sim \mathcal{N}(\mu,\tau^2).
\]
The model also has a global prior on $\mu$, $\tau^2$, and $\sigma_s^2$.
The state of the MCMC chain at iteration~$t$ is $x_t=(\theta_{1:S,t}, \sigma^2_{1:S,t}, \mu_t, \tau_t^2) \in \mathbb{R}^{2S + 2}$. A Gibbs sampler iteratively updates each state variable by sampling from its conditional distribution.
Fixing the random numbers used in the conditional updates turns the sampler into a deterministic input-driven state-space model, $x_{t+1}=f_t(x_t)$, where one transition corresponds to a full Gibbs sweep. Details are given in Appendix~\ref{app:gibbs}.

This model provides a setting where the useful Jacobian structure is low-rank rather than diagonal. Cross-school coupling is mediated primarily by the two global variables $\mu$ and $\tau^2$, which aggregate information from the school-level variables and broadcast it back through the conditional updates. Thus, the dominant off-diagonal structure has low effective rank, with rank two
providing a natural approximation. Meanwhile, Gibbs updates resample variables from their conditional distributions, so the diagonal entries of the dynamics Jacobian are small (Appendix~\ref{app:gibbs}). Motivated by this structure, we set the diagonal term to zero for Parallel Broyden variants and use rank $m=2$, matching the two global variables $(\mu,\tau^2)$. In this purely low-rank setting, no rank compression
is required.

\begin{figure}
    \centering
    \includegraphics[width=1.\linewidth]{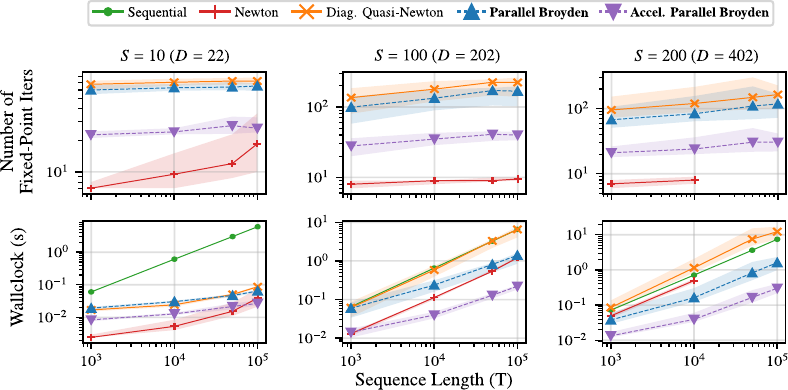}
    \caption{
\textbf{Hierarchical Gibbs sampling.}
Across 32 problem instances, we report median iteration counts and
runtimes with interquartile ranges as we vary trajectory length $T$ and state dimension $D=2S+2$. Parallel Broyden variants require fewer iterations
than diagonal quasi-Newton by capturing low-rank hierarchical coupling.
Accelerated Parallel Broyden is the fastest approximate method in every setting, and the fastest overall except for small problems (small $S$ or short $T$). It also remains
applicable at the largest settings, where exact Newton runs out of memory.
}
    \label{fig:gibbs}
\end{figure}

We evaluate all methods on $32$ random problem instances, each with independently sampled observations and initialization, as we vary the trajectory length $T$ and state dimension $D=2S+2$ (Figure~\ref{fig:gibbs}). The two baselines fail in opposite ways. Exact Newton captures the coupling and uses the fewest iterations, but its dense Jacobian cost grows cubically with $D$, and it runs out of memory at the largest settings. Diagonal quasi-Newton neglects the dominant hierarchical coupling, so it requires many more iterations, and since obtaining its exact diagonal still requires forming the full $D \times D$ Jacobian, at $S \geq 100$ it is no faster than sequential evaluation. This cost can be avoided with the Hutchinson estimator~\citep{hutchinson1989stochastic}, which estimates the diagonal from a few Jacobian-vector products without forming the Jacobian, as in recent quasi-Newton approaches for parallel MCMC~\citep{zoltowskiparallelizing}. Here, however, this estimate causes the iterations to fail to converge (Appendix~\ref{app:gibbs}).

Both Parallel Broyden variants need fewer iterations than diagonal quasi-Newton in every setting, indicating that the low-rank secant corrections capture the cross-school coupling. Accelerated Parallel Broyden needs $2.3$--$4.2\times$ fewer iterations still than Parallel Broyden and is the fastest approximate method throughout. Exact Newton remains fastest only for small problems (e.g., $S=10$ with $T \leq 50\mathrm{K}$). Elsewhere, Accelerated Parallel Broyden is fastest overall, with an advantage that grows with $T$ and $D$. At $S=100$ and $T=100\mathrm{K}$, it is $5.5\times$ faster than exact Newton and $30\times$ faster than sequential evaluation. At $S=200$, both Broyden variants beat every baseline at every $T$ and remain applicable after exact Newton runs out of memory. At $T=100\mathrm{K}$, Accelerated Parallel Broyden is $24.7\times$ faster than sequential evaluation and $41\times$ faster than diagonal quasi-Newton.


\subsection{Denoising diffusion models}

We finally consider parallelizing diffusion sampling with a pretrained DiT image
denoiser, a transformer-based diffusion model over image patches
\citep{peebles2023scalable}. We use DDIM sampling~\citep{song2020denoising},
which evolves a noisy latent toward a clean image through deterministic denoising
updates
\[
    s_{t-1}
    =
    c_x(t)s_t + c_\epsilon(t)\epsilon_\theta(s_t,t).
\]
Re-indexing in forward time, we write this as a time-indexed state-space model
$x_\ell=f_\ell(x_{\ell-1})$ with $t=T-\ell+1$. The state dimension is $D=4096$. The main challenge is Jacobian evaluation. The local dynamics Jacobian block is $A_\ell(x_{\ell-1})
    =
    c_x(t)I
    +
    c_\epsilon(t)
    \frac{\partial \epsilon_\theta}{\partial x}(x_{\ell-1},t),$
so exact Newton would require differentiating through the full DiT denoiser at
each time step, which is prohibitively expensive at this scale. Since $A_\ell$ is moreover dominated by its diagonal (Appendix~\ref{app:ddim_jac}), this benchmark tests whether a good approximation can be obtained without differentiating through the denoiser, rather than whether a richer one reduces iterations.

We compare against two recent parallel diffusion samplers. The first is Picard iteration~\citep{shih2023parallel}, which corresponds in our formulation to using the identity local approximation $\widetilde A_\ell=I_D$ \citep{gonzalez2025unifying}. The second is ParaTAA~\citep{tang2024accelerating}, which uses multisecant conditions to approximate the full inverse residual Jacobian $\mathbf{J}^{-1}$. In contrast, Parallel Broyden approximates the local dynamics blocks $A_\ell$ used in the scan. Implementation details and hyperparameter searches for all methods are given in Appendix~\ref{app:ddim}.

Figure~\ref{fig:fig3} reports iteration counts and wall-clock runtimes for 50-step DDIM sampling with the pretrained DiT model, averaged over $32$ images. Parallel Broyden and Accelerated Parallel Broyden converge in roughly the same number of outer iterations ($14.9$ on average), and diagonal quasi-Newton requires only slightly more ($16.8$), consistent with $A_\ell$ being dominated by its diagonal. All three local dynamics-Jacobian methods require fewer iterations than Picard and ParaTAA, with Parallel Broyden reducing the mean iteration count by $1.21\times$ relative to Picard and by $1.68\times$ relative to ParaTAA. This supports approximating local dynamics Jacobian blocks directly rather than using either the identity local approximation or the inverse Jacobian of the full residual system.

The runtime comparison highlights the cost of obtaining the Jacobian
approximation. Diagonal quasi-Newton estimates the diagonal of $A_\ell$ with a
single-probe Hutchinson estimator, which still requires one Jacobian-vector
product through the denoiser per time step and iteration, making it the slowest
parallel method. Picard iteration avoids differentiation but needs more
iterations. Parallel Broyden avoids both costs and achieves the fastest
parallel runtime: $3.32\times$ faster than diagonal quasi-Newton,
$1.12\times$ faster than Picard, and $1.53\times$ faster than ParaTAA, with
Accelerated Parallel Broyden performing similarly. Its advantage here therefore
comes from avoiding automatic differentiation rather than from a richer
Jacobian approximation. Upon convergence, all methods produce images that are
visually indistinguishable from the sequential reference
(Appendix~\ref{app:ddim_quality}).

All parallel methods remain slower than sequential DDIM on a single H100 GPU,
because evaluating the DiT denoiser across many timesteps in parallel quickly
becomes FLOP-bound. Prior work therefore uses multiple GPUs to amortize these
evaluations \citep{shih2023parallel,tang2024accelerating}, where fewer outer
iterations directly mean fewer rounds of parallel denoiser evaluations.

\begin{figure}
    \centering
    \includegraphics[width=1.\linewidth]{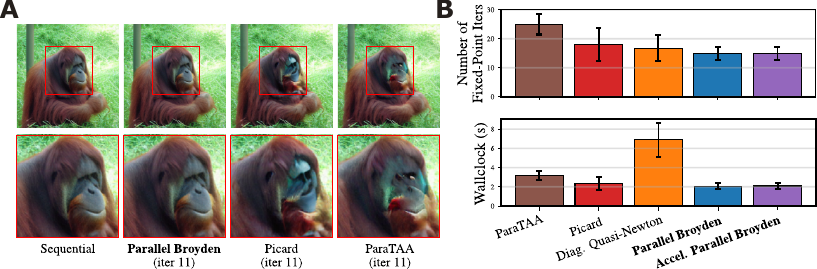}
    \caption{
\textbf{DDIM sampling with a pretrained DiT denoiser.}
\textbf{A:} After $11$ iterations, Parallel Broyden is visually
indistinguishable from the sequential reference, while ParaTAA and Picard have
not fully converged (bottom: zoom on boxed region). \textbf{B:} Mean iteration
counts and runtimes over $32$ images ($\pm$ s.d.). Local dynamics-Jacobian
methods need fewer iterations than Picard and ParaTAA, and Parallel Broyden variants are
the fastest parallel methods by avoiding differentiation through the denoiser.
}
    \label{fig:fig3}
\end{figure}

\section{Conclusion}

We introduced Parallel Broyden, a family of Broyden-style quasi-Newton methods that build diagonal-plus-low-rank approximations of local dynamics Jacobian blocks from trajectory secants. By avoiding automatic differentiation while remaining compatible with parallel scan through rank compression, Parallel Broyden provides a more scalable alternative to exact parallel Newton in high dimensions and a richer approximation class than diagonal quasi-Newton methods. We proved finite-step recovery under the compressed scan, and showed that convergence is tracked by the compression error and by the block approximation error along the current trajectory error, rather than in spectral norm.

In our experiments, Parallel Broyden reduces iterations relative to diagonal
quasi-Newton when coupling is approximately low rank, and improves runtime when
dense Jacobian operations or automatic differentiation are costly. The multisecant Accelerated Parallel Broyden variant is often strongest,
consistent with its observed smaller block error along the current error near
convergence.

Several limitations remain. Parallel Broyden is designed for dynamics whose dominant off-diagonal Jacobian structure can be captured with a small number of low-rank corrections. When accurate approximations require large rank, the advantage diminishes, as
scan and compression costs grow with rank. Moreover, our convergence analysis is local and conservative. Relating block error along the current trajectory error to the secant updates is a natural next step. Beyond the settings studied here, DPLR structure also arises when a few global variables mediate interactions among many local ones, as in mean-field coupled oscillators, and Jacobian-free methods suit neural ODEs and black-box simulators, where Jacobians are costly or unavailable. These are promising applications for future work. More broadly, the compressed approximate scan applies to any DPLR block approximation, including autodiff-based sketches when differentiation is affordable.

\newpage
\section*{Acknowledgments}
We thank Xavier Gonzalez and the other members of the Linderman Lab for helpful feedback throughout this project, and Leo Kozachkov for valuable insights into the asymptotic theory of Newton's method. S.W.L. is supported by grants from the NIH (U01NS136507, R01NS131987,
RF1MH133778, R01AG097491, \& R01NS130789), the NSF (2440859), and the Simons and McKnight Foundations.

\bibliographystyle{unsrtnat}

\bibliography{reference}

\newpage
\appendix

\section{Overview of Parallel Broyden}
\label{app:overview}

\begin{figure}[h]
    \centering
    \includegraphics[width=1\linewidth]{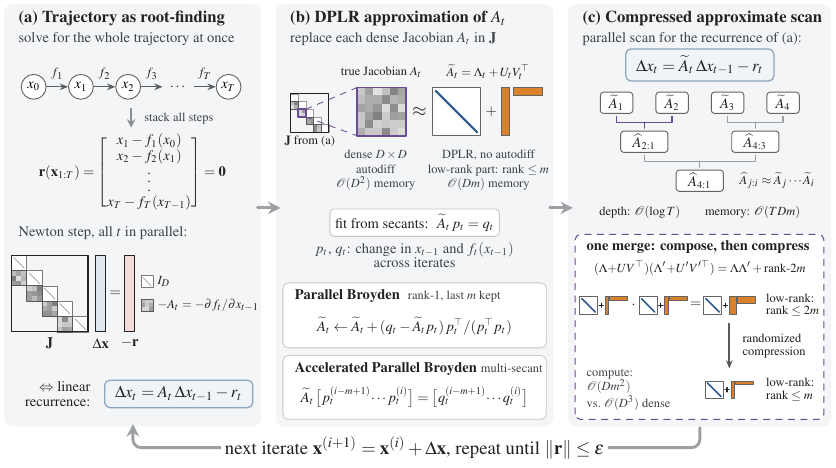}
    \caption{\textbf{Overview of Parallel Broyden.}
\textbf{(a)} Evaluating the sequence is cast as solving $\mathbf r(\mathbf x_{1:T}) = \mathbf 0$ with $r_t = x_t - f_t(x_{t-1})$. Each Newton step solves a block-bidiagonal system $\mathbf J\,\Delta\mathbf x = -\mathbf r$, which is equivalent to the linear recurrence $\Delta x_t = A_t\,\Delta x_{t-1} - r_t$ with Jacobians $A_t = \partial f_t/\partial x_{t-1}$.
\textbf{(b)} Each dense Jacobian $A_t$ is replaced by a diagonal-plus-low-rank (DPLR) matrix $\widetilde A_t = \Lambda_t + U_tV_t^\top$, with $\Lambda_t$ diagonal and $U_t, V_t \in \mathbb{R}^{D\times m}$. It is fitted without automatic differentiation from secants $p_t = x_{t-1}^{(i+1)} - x_{t-1}^{(i)}$ and $q_t = f_t(x_{t-1}^{(i+1)}) - f_t(x_{t-1}^{(i)})$, either by rank-one Broyden updates (Parallel Broyden) or by fitting the last $m$ secants jointly (Accelerated Parallel Broyden).
\textbf{(c)} The recurrence is solved with a parallel scan over the $\widetilde A_t$. Each merge multiplies two DPLR matrices, which raises the rank of the low-rank part to at most $2m$, then compresses it back to rank $m$ with randomized compression, so $\widehat A_{j:i} \approx \widetilde A_j \cdots \widetilde A_i$. Each merge costs $\mathcal O(Dm^2)$ instead of $\mathcal O(D^3)$, giving $\mathcal O(\log T)$ depth and $\mathcal O(TDm)$ memory.
The iterate is updated as $\mathbf x^{(i+1)} = \mathbf x^{(i)} + \Delta\mathbf x$ until $\|\mathbf r\| \le \epsilon$.}
\label{fig:method}
    \label{fig:overview}
\end{figure}

\section{Effect of rank compression}
\label{app:compression}

Our compressed approximate scan keeps diagonal-plus-low-rank (DPLR) factors at
fixed rank by compressing the low-rank correction after each scan merge. This
compression is necessary to prevent rank growth in long trajectories, but it
also raises two questions. First, does compression preserve the finite-step
trajectory recovery property of scan-based Newton methods? Second, how does the
chosen compression rank affect convergence and runtime in practice?

We address these questions in three steps. First, we prove that finite-step
trajectory recovery is preserved under the compressed approximate scan. Second, we describe two compression schemes: an exact small-SVD compression and a randomized factorized compression that avoids forming
dense matrices. Third, we empirically study how the compression rank affects
convergence and runtime on the low-rank RNN benchmark.

\subsection{Proof of Theorem \ref{thm:global_recovery}}
\label{app:global_recovery}

\renewcommand{\thereptheorem}{\ref{thm:global_recovery}}
\begin{reptheorem}[Global convergence with approximate scan]
Starting from any initial trajectory $\mathbf{x}^{(0)}_{1:T}$, after $i$
iterations using the approximate parallel scan with low-rank compression, we
have $x_t^{(i)} = x_t^\star$ for all $t \leq i$. In particular, the method
recovers the exact trajectory $\mathbf{x}^\star_{1:T}$ in at most $T$
iterations.
\end{reptheorem}

\begin{proof}
\textbf{Recursion.}
At iteration $i$, let $\widetilde A_t^{(i)}$ denote the approximate dynamics
Jacobian block used in place of
\[
    A_t^{(i)}
    =
    \frac{\partial f_t}{\partial x_{t-1}}
    \left(x_{t-1}^{(i)}\right).
\]
The approximate Newton update solves the linear recurrence
\begin{equation}
    \Delta x_1^{(i+1)}
    =
    -r_1(\mathbf{x}^{(i)}_{1:T}),
    \qquad
    \Delta x_t^{(i+1)}
    =
    \widetilde A_t^{(i)} \Delta x_{t-1}^{(i+1)}
    -
    r_t(\mathbf{x}^{(i)}_{1:T}),
    \quad t>1.
    \label{eq:recurrence}
\end{equation}

\textbf{Compressed scan.}
Each position $t$ is represented by an affine summary
\[
    (M_t,c_t),
    \qquad
    M_t=\widetilde A_t^{(i)}, \qquad
    c_t=-r_t(\mathbf{x}^{(i)}_{1:T}),
\]
corresponding to the local update rule
\[
    x \mapsto M_t x + c_t .
\]
The compressed scan combines two adjacent summaries $(M_a,c_a)$ and
$(M_b,c_b)$ by
\begin{equation}
    (M_b,c_b)\star(M_a,c_a)
    :=
    \left(
        \Comp{M_bM_a},
        M_b c_a + c_b
    \right),
    \label{eq:compressed_merge}
\end{equation}
where $\Comp{\cdot}$ denotes low-rank compression of the matrix part. Because of
compression, this merge need not equal exact affine composition and need not be
associative. The only property we use is that compression is applied only to the
matrix component, while the offset component is updated by
\[
    c_{\mathrm{new}} = M_b c_a + c_b.
\]

The scan returns, for each prefix $1{:}t$, an affine summary
\[
    (M_{1:t},c_{1:t}),
\]
obtained from the leaf summaries by a binary tree of merges
\eqref{eq:compressed_merge}, each combining the summaries of two adjacent
segments. This holds for standard scan algorithms, and we make no other
assumption on the shape of the tree. The update is recovered from the
returned offset:
\begin{equation}
    \Delta x_t^{(i+1)} = c_{1:t}.
    \label{eq:scan_output}
\end{equation}

\textbf{Induction.}
We prove by induction on $i$ that
\[
    x_t^{(i)} = x_t^\star
    \qquad
    \text{for all } t \le i.
\]

\textit{Initialization.}
Since $x_0$ and $f_1$ are fixed, $x_1^\star=f_1(x_0)$. The prefix at position
$1$ is a leaf, so no merge or compression is applied and
\[
    \Delta x_1^{(1)}
    =
    c_1
    =
    -r_1(\mathbf{x}^{(0)}_{1:T})
    =
    -\bigl(x_1^{(0)}-f_1(x_0)\bigr).
\]
Therefore
\[
    x_1^{(1)}
    =
    x_1^{(0)}+\Delta x_1^{(1)}
    =
    f_1(x_0)
    =
    x_1^\star.
\]

\textit{Inductive step.}
Assume that
\[
    x_t^{(i)}=x_t^\star
    \qquad
    \text{for all } t\le i.
\]
Then for every $t\le i$,
\[
    r_t(\mathbf{x}^{(i)}_{1:T})
    =
    x_t^{(i)}-f_t(x_{t-1}^{(i)})
    =
    x_t^\star-f_t(x_{t-1}^\star)
    =
    0.
\]
Hence the corresponding leaf offsets satisfy
\[
    c_t=-r_t(\mathbf{x}^{(i)}_{1:T})=0
    \qquad
    \text{for all } t\le i.
\]

Now consider any prefix contained entirely in $[1,i]$. Every leaf in this prefix
has zero offset. Moreover, if two summaries have zero offsets, then their
compressed merge also has zero offset:
\[
    c_a=0,\quad c_b=0
    \quad\Longrightarrow\quad
    c_{\mathrm{new}}
    =
    M_b c_a+c_b
    =
    0,
\]
regardless of the compressed matrix part. By induction over the scan tree,
every prefix summary over $[1,t]$ with $t\le i$ therefore has offset
\[
    c_{1:t}=0.
\]
Using \eqref{eq:scan_output}, this gives
\[
    \Delta x_t^{(i+1)}=0
    \qquad
    \text{for all } t\le i,
\]
and hence
\[
    x_t^{(i+1)}=x_t^{(i)}=x_t^\star
    \qquad
    \text{for all } t\le i.
\]

It remains to show that state $i+1$ is recovered. The scan may build the
prefix $1{:}i+1$ from segments of any shape, so we prove a claim about every
segment that the scan forms inside $[1,i+1]$.

\emph{Claim.} Let $(M,c)$ be the summary that the scan forms for a segment
$\{p,\dots,q\}$ with $q\le i+1$. Then $c=0$ if $q\le i$, and $c=c_{i+1}$ if
$q=i+1$.

\emph{Proof of the claim.} By induction over the scan tree. For a leaf
$\{q\}$, the offset is $c_q$, which is $0$ if $q\le i$ and $c_{i+1}$ if
$q=i+1$. Now consider a merge of a left segment $a=\{p,\dots,m\}$ and a right
segment $b=\{m{+}1,\dots,q\}$. Since $m<q\le i+1$, we have $m\le i$, so
$c_a=0$ by the inductive hypothesis. Hence
\[
    c_{\mathrm{new}}=M_b c_a+c_b=c_b ,
\]
whatever the compressed matrix $M_b$ is. Segment $b$ ends at $q$, so $c_b$,
and therefore $c_{\mathrm{new}}$, is as claimed.

Applying the claim to the prefix $1{:}i+1$ gives
\[
    c_{1:i+1}=c_{i+1}=-r_{i+1}(\mathbf{x}^{(i)}_{1:T}),
\]
so, by \eqref{eq:scan_output},
\[
    \Delta x_{i+1}^{(i+1)}
    =
    -\bigl(x_{i+1}^{(i)}-f_{i+1}(x_i^{(i)})\bigr).
\]
By the induction hypothesis, $x_i^{(i)}=x_i^\star$, so
\[
    \Delta x_{i+1}^{(i+1)}
    =
    -\bigl(x_{i+1}^{(i)}-f_{i+1}(x_i^\star)\bigr)
    =
    x_{i+1}^\star-x_{i+1}^{(i)}.
\]
Thus
\[
    x_{i+1}^{(i+1)}
    =
    x_{i+1}^{(i)}+\Delta x_{i+1}^{(i+1)}
    =
    x_{i+1}^\star.
\]

Therefore, after iteration $i+1$, the first $i+1$ states are exact. By
induction, after $i$ iterations we have $x_t^{(i)}=x_t^\star$ for all $t\le i$,
and the full trajectory is recovered in at most $T$ iterations.
\end{proof}

\subsection{Compression schemes}
\label{app:compression_schemes}

During the compressed approximate scan, composing two rank-at-most $m$ DPLR blocks
produces a new DPLR block whose low-rank correction has rank at most $2m$.
Thus, after each merge, we need to compress a factorized correction
\[
    \Delta = U V^\top,
    \qquad
    U,V \in \mathbb{R}^{D\times r},
    \qquad
    r \leq 2m,
\]
back to rank at most $m$. We consider two compression schemes: an exact
best-rank-$m$ compression based on a small SVD, and a randomized factorized
compression that avoids computing an SVD.

\paragraph{Exact compression via a small SVD.}
The exact compression computes the best rank-$m$ approximation to
$\Delta=UV^\top$ in Frobenius norm. By the Eckart--Young--Mirsky theorem \citep{eckart1936approximation, mirsky1960symmetric}, this
approximation is obtained by truncating the singular value decomposition of
$\Delta$. Since $\Delta$ has rank at most $r\leq 2m$, this SVD can be computed
without forming the dense $D\times D$ matrix.

Let
\[
    U = Q_U R_U,
    \qquad
    V = Q_V R_V
\]
be thin QR decompositions, where
\[
    Q_U,Q_V \in \mathbb{R}^{D\times r},
    \qquad
    R_U,R_V \in \mathbb{R}^{r\times r}.
\]
Then
\[
    \Delta
    =
    U V^\top
    =
    Q_U R_U R_V^\top Q_V^\top.
\]
Define the small core matrix
\[
    C := R_U R_V^\top \in \mathbb{R}^{r\times r}.
\]
We compute its SVD
\[
    C = \bar U \Sigma \bar V^\top.
\]
Since $Q_U$ and $Q_V$ have orthonormal columns, this gives an SVD of the
rank-$r$ correction:
\[
    \Delta
    =
    (Q_U\bar U)\Sigma(Q_V\bar V)^\top.
\]
Let $\bar U_m,\bar V_m$ denote the first $m$ singular vectors and let
$\Sigma_m$ contain the top $m$ singular values. The best rank-$m$ approximation
is therefore
\[
    \Delta_m
    =
    (Q_U\bar U_m)\Sigma_m(Q_V\bar V_m)^\top.
\]
Equivalently, we store it in factored form as
\[
    \Delta_m
    =
    \widetilde U \widetilde V^\top,
    \qquad
    \widetilde U := Q_U\bar U_m\Sigma_m,
    \qquad
    \widetilde V := Q_V\bar V_m .
\]
This compression is optimal among all rank-$m$ approximations:
\[
    \Delta_m
    =
    \arg\min_{\operatorname{rank}(B)\leq m}
    \|\Delta-B\|_F .
\]

The cost is dominated by the two thin QR decompositions and the small SVD.
For $r\leq 2m$, the QR decompositions cost $\mathcal{O}(D r^2)$, the core SVD
costs $\mathcal{O}(r^3)$, and forming the compressed factors costs
$\mathcal{O}(D r m)$. Thus the total cost is
\[
    \mathcal{O}(D m^2 + m^3),
\]
with memory $\mathcal{O}(Dm)$ for the compressed factors.

\paragraph{Randomized factorized compression.}
The randomized compression avoids the small SVD and compresses directly in
factorized form. Given
\[
    \Delta = U V^\top,
    \qquad
    U,V\in\mathbb{R}^{D\times r},
    \qquad
    r\leq 2m,
\]
we draw a Gaussian test matrix
\[
    \Omega \in \mathbb{R}^{D\times m}
\]
and form the sketch
\[
    Y
    =
    \Delta \Omega
    =
    U(V^\top \Omega)
    \in \mathbb{R}^{D\times m}.
\]
We then compute a thin QR decomposition
\[
    Y = QR,
    \qquad
    Q\in\mathbb{R}^{D\times m},
\]
so that the columns of $Q$ approximate the dominant column space of
$\Delta$. Projecting $\Delta$ onto this subspace gives
\[
    \Delta
    \approx
    QQ^\top \Delta
    =
    Q(Q^\top U)V^\top.
\]
This can again be written as a rank-$m$ factorization
\[
    QQ^\top \Delta
    =
    \widetilde U \widetilde V^\top,
    \qquad
    \widetilde U := Q,
    \qquad
    \widetilde V := V(U^\top Q).
\]
Indeed,
\[
    \widetilde U\widetilde V^\top
    =
    Q\bigl(V(U^\top Q)\bigr)^\top
    =
    Q(Q^\top U)V^\top
    =
    QQ^\top U V^\top .
\]

The randomized scheme is not guaranteed to produce the exact best rank-$m$
approximation, but it avoids the SVD and preserves the factorized
representation throughout. Its cost is dominated by multiplying the factors
against the sketch and orthonormalizing $Y$. Since $r\leq 2m$, forming
$V^\top\Omega$, forming $Y=U(V^\top\Omega)$, computing the QR factorization of
$Y$, and forming $\widetilde V=V(U^\top Q)$ together cost
\[
    \mathcal{O}(D m^2),
\]
up to lower-order $\mathcal{O}(m^3)$ terms from the QR decomposition. The memory
cost is $\mathcal{O}(Dm)$ for the compressed factors.

In the experiments, we use randomized compression by default because it is
simple, factorized, and avoids dense matrix construction. The exact
small-SVD compression provides a useful reference implementation and gives the
optimal rank-$m$ correction after each merge.

\subsection{Empirical impact of compression}
\label{app:compression_impact}

We study the effect of rank compression on the low-rank RNN benchmark using
Accelerated Parallel Broyden. This experiment isolates the cost of solving the
block-bidiagonal Newton system: all methods use the same multisecant Broyden
block approximations and differ only in how the linearized recurrence is solved.
We compare three variants. The first solves the recurrence by exact sequential
forward substitution, which introduces no scan compression error but has linear
depth in $T$. The second uses the compressed parallel scan with exact QR+SVD rank
compression after each merge. The third uses the randomized factorized
compression described above. We report median iteration counts and runtimes with
interquartile ranges over $32$ random realizations.

\begin{figure}
    \centering
    \includegraphics[width=\linewidth]{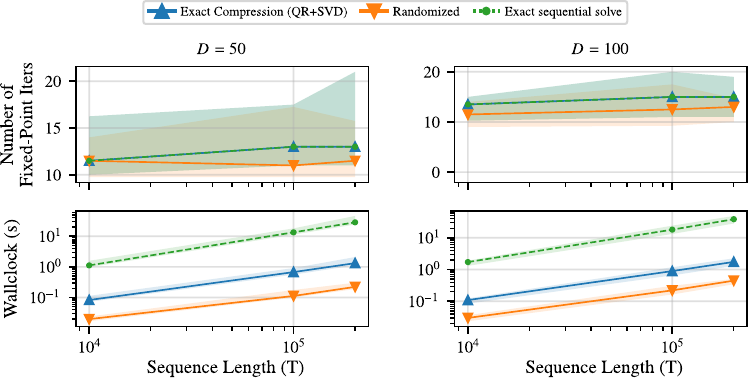}
    \caption{\textbf{Effect of compression on the low-rank RNN benchmark.} We compare exact sequential forward substitution,
    compressed scan with exact QR+SVD rank compression, and compressed scan with
    randomized factorized compression. Points show medians over $32$ random
    realizations, with error bars indicating interquartile ranges. All three
    variants require similar numbers of Newton iterations, indicating that
    compression does not materially degrade convergence. Randomized compression
    gives the lowest runtime by avoiding both sequential depth and the small
    SVDs used by exact compression.}
    \label{fig:compression_impact}
\end{figure}

Figure~\ref{fig:compression_impact} shows that compression has little effect on
iteration counts. Exact sequential solution and QR+SVD compression have
identical median iteration counts in all tested settings, confirming that the
compressed scan preserves the relevant low-rank structure in this benchmark. The
randomized compression has comparable, and sometimes slightly lower, iteration
counts, with medians between roughly $11$ and $13$ iterations across all
dimensions and horizons.

The main difference is runtime. Exact sequential forward substitution is much
slower because it solves the linearized recurrence with linear depth in $T$.
At $D=100$ and $T=200\mathrm{K}$, exact QR+SVD compression is $22.0\times$
faster than the exact sequential solve, and randomized compression is
$85.6\times$ faster. Randomized compression is also $3.9\times$ faster than
exact QR+SVD compression in this setting. Thus, in this experiment, rank
compression enables the scan to retain the convergence behavior of the exact
linear solve while giving substantial runtime gains, and the randomized scheme
provides the best practical tradeoff.

\subsection{Choice of parallel scan tree}
\label{app:scan_tree}

The parallel scan composes the affine maps $x\mapsto \widetilde A_t x + c_t$ along a binary tree. With exact composition, the operator is associative and every tree gives the same result. With DPLR blocks, however, each composition of two rank-$m$ blocks is compressed back to rank $m$ with a randomized sketch, so the combine operation is only approximately associative, and the result depends on which products are formed and compressed. We compare two standard trees. The Blelloch scan~\citep{blelloch1989scans} is work-efficient. An up-sweep builds the products of contiguous blocks of $2^k$ steps, and a down-sweep distributes them to all prefixes, for $O(T)$ compositions and depth about $2\log_2 T$. The Hillis--Steele scan~\citep{hillis1986data} instead updates every prefix at each of its $\log_2 T$ levels by composing it with the prefix $2^k$ steps earlier, for $O(T\log T)$ compositions and depth $\log_2 T$. The two trees therefore differ in the number and order of compressions applied to each prefix product. All other experiments in this paper use the Blelloch-type scan implemented by \texttt{jax.lax.associative\_scan}.

To check that our results do not depend on this choice, we ran both variants of Parallel Broyden on the low-rank RNNs with the settings of Appendix~\ref{app:low_rank_rnn} and both scan trees, in \texttt{float64} since only iteration counts are compared. The two trees also draw different random sketches, so the comparison includes both effects. The choice of tree has no meaningful effect on convergence (Table~\ref{tab:scan_tree}). Median iteration counts agree in $21$ of $24$ settings and differ by at most half an iteration otherwise, and the number of converged chains is identical in every setting. At the level of individual chains, $80\%$ of chains take exactly the same number of iterations under both trees, and among chains that converge under both, $94\%$ are within $5$ iterations. The mean relative difference of $3.7\%$ comes from a few chains close to the convergence boundary.

\begin{table}[h]
    \centering
    \caption{\textbf{Effect of the scan tree on iteration counts} (low-rank RNNs, $D=200$, \texttt{float64}). Median iterations over $32$ chains with the Blelloch (B) and Hillis--Steele (HS) scans, with non-converged chains counted at the $1000$-iteration cap.}
    \label{tab:scan_tree}
    \begin{tabular}{llrrrrr}
        \toprule
        & & & \multicolumn{2}{c}{PB} & \multicolumn{2}{c}{Accel.\ PB} \\
        \cmidrule(lr){4-5} \cmidrule(lr){6-7}
        Model & $\Delta t$ & $T$ & B & HS & B & HS \\
        \midrule
        Standard & $0.5$ & $10^3$ & $13$   & $13$   & $9$    & $9$    \\
                 &       & $10^4$ & $15$   & $15$   & $11$   & $11$   \\
                 &       & $10^5$ & $14$   & $14$   & $12.5$ & $13$   \\
                 & $3$   & $10^3$ & $25$   & $25$   & $11$   & $11$   \\
                 &       & $10^4$ & $26$   & $26$   & $12$   & $12$   \\
                 &       & $10^5$ & $26$   & $26$   & $13.5$ & $13.5$ \\
        \midrule
        MLP      & $0.5$ & $10^3$ & $13$   & $13$   & $10$   & $10$   \\
                 &       & $10^4$ & $14$   & $14$   & $10.5$ & $11$   \\
                 &       & $10^5$ & $14.5$ & $14.5$ & $12$   & $12$   \\
                 & $3$   & $10^3$ & $25$   & $25$   & $12$   & $12$   \\
                 &       & $10^4$ & $24$   & $24$   & $12$   & $12$   \\
                 &       & $10^5$ & $24.5$ & $24.5$ & $12$   & $11.5$ \\
        \bottomrule
    \end{tabular}
\end{table}

\section{Convergence analysis of quasi-Newton updates with compressed scan}
\label{app:conv}

This appendix proves Theorem~\ref{thm:cvg_bound}. We first fix notation and
state the assumptions precisely, and give a sufficient condition in terms of
the true dynamics (\ref{app:conv_setup}). We then derive a one-step error bound
for any scan of the form \eqref{eq:compressed_merge}, which extends
Proposition~3 of \citet{gonzalez2025unifying} to compressed scans
(\ref{app:conv_onestep}). Next, we bound the compressed products used by the
scan, which proves part~(i) of the theorem (\ref{app:conv_products}). We then
prove parts~(ii) and~(iii) in a general form that does not require
$\lambda_c<0$ (\ref{app:conv_proof}), and discuss consequences and limitations
(\ref{app:conv_consequences}). Empirical measurements of the quantities in the
theorem are reported in Appendix~\ref{app:conv_diagnostics}.

\paragraph{Proof idea.}
The analysis follows Proposition~3 of \citet{gonzalez2025unifying}, which
writes the error after an approximate Newton step as $\widetilde{\mathbf J}^{-1}$
applied to the block error $(\widetilde{\mathbf J}-\mathbf J)e$ plus a
second-order remainder. The key observation is that compression acts only on
the matrix components of the scan elements, which do not depend on the
residual, so the compressed step remains linear in the residual,
$\Delta\mathbf x=-\mathbf P\,\mathbf r$, with $\mathbf P$ playing the role of
$\widetilde{\mathbf J}^{-1}$ (Lemma~\ref{lem:scan_linear}). The one-step
argument therefore carries over, with one additional term $\Gamma e$, where
$\Gamma=I-\mathbf P\widetilde{\mathbf J}$ vanishes without compression
(Lemma~\ref{lem:gamma}). Unlike Proposition~3, which bounds the block error by
$\|\widetilde{\mathbf J}-\mathbf J\|_2$, we keep it applied to the current
trajectory error, which gives $\rho$. The main new ingredient is bounding
$\mathbf P$ and $\Gamma$: compressed products are not products of the blocks,
so we expand them over the scan tree and control the relative compression
error introduced at each merge (Lemma~\ref{lem:comp_products}).

\subsection{Setting, notation, and assumptions}
\label{app:conv_setup}

\paragraph{Iterates and errors.}
At iteration $i$, write $A_t^{(i)}:=A_t(x_{t-1}^{(i)})$ for the exact dynamics
Jacobian block and $\widetilde A_t^{(i)}$ for the approximate block used by the
solver (for Parallel Broyden, the damped block of
Appendix~\ref{app:broyden_damping}). The trajectory error is
$e^{(i)}:=\mathbf x^{(i)}_{1:T}-\mathbf x^\star_{1:T}$, with blocks
$e_t^{(i)}=x_t^{(i)}-x_t^\star\in\mathbb R^D$, and $e_0^{(i)}=0$ since $x_0$ is
fixed. We write $\mathbf J^{(i)}:=\mathbf J(\mathbf x^{(i)}_{1:T})$ and
$\widetilde{\mathbf J}^{(i)}$ for the exact and approximate residual Jacobians.
When the iteration index is clear, we drop the superscript $(i)$. We analyze
full steps $\mathbf x^{(i+1)}=\mathbf x^{(i)}+\Delta\mathbf x^{(i+1)}$. See
Appendix~\ref{app:conv_consequences} for damped steps.

\paragraph{Per-step norm.}
For $v=(v_1,\dots,v_T)\in\mathbb R^{TD}$ with $v_t\in\mathbb R^D$, define
\[
    \|v\|_\infty:=\max_{1\le t\le T}\|v_t\|_2 .
\]
For a block matrix $M\in\mathbb R^{TD\times TD}$ with blocks
$M_{t,s}\in\mathbb R^{D\times D}$, let
$\|M\|_\infty:=\sup_{v\neq0}\|Mv\|_\infty/\|v\|_\infty$ be the induced norm.
Since $\|(Mv)_t\|_2\le\sum_s\|M_{t,s}\|_2\|v_s\|_2$,
\begin{equation}
    \|M\|_\infty\le\max_{t}\sum_{s}\|M_{t,s}\|_2 .
    \label{eq:app_rowsum}
\end{equation}
We use this norm rather than the Euclidean norm on $\mathbb R^{TD}$ used by
\citet{gonzalez2025unifying}: for bounded states, $\|e^{(0)}\|_\infty$ does not
grow with $T$, whereas $\|e^{(0)}\|_2$ typically grows like $\sqrt T$, which
would make the region of convergence shrink artificially with $T$.

\paragraph{Geometric sums.}
For $\mu\in\mathbb R$, define
\[
    S_T(\mu):=\sum_{n=0}^{T-1}e^{\mu n},
    \qquad\text{so that}\qquad
    S_T(\mu)\le\frac{1}{1-e^{\mu}}\ \text{ if }\mu<0,
    \qquad S_T(0)=T .
\]
$S_T$ is increasing in $\mu$, and for $\mu>0$ it grows like $e^{\mu T}$.

\paragraph{Compressed scan.}
As in Appendix~\ref{app:global_recovery}, position $t$ is represented by the
leaf summary $(\widetilde A_t,\,c_t)$ with $c_t=-r_t(\mathbf x^{(i)}_{1:T})$,
summaries of adjacent segments are combined by the merge
\eqref{eq:compressed_merge},
\[
    (M_b,c_b)\star(M_a,c_a)=\big(\Comp{M_bM_a},\,M_bc_a+c_b\big),
\]
and the update is read off the prefix offsets, $\Delta x_t^{(i+1)}=c_{1:t}$.
We assume that each prefix summary returned by the scan is obtained from the
leaf summaries by a binary tree of such merges, each combining the summaries of
two adjacent segments. This holds for standard scan algorithms. We call each
merge a \emph{node} $\nu$. It covers a segment of consecutive time steps
$\{j{+}1,\dots,k\}$ of length $|\nu|=k-j$. At node $\nu$, the scan forms
$X_\nu:=\widehat M_{\mathrm{right}}\widehat M_{\mathrm{left}}$ from the stored
matrices of its two children and stores
\[
    \widehat M_\nu:=\Comp{X_\nu}=X_\nu+\Delta_\nu,
    \qquad
    \Delta_\nu:=\Comp{X_\nu}-X_\nu .
\]
Leaves are never compressed: the stored matrix of the leaf at $t$ is
$\widetilde A_t$. We write
$\Pi_{t,s}:=\widetilde A_t\widetilde A_{t-1}\cdots\widetilde A_{s+1}$ for the
exact product over $\{s{+}1,\dots,t\}$, with $\Pi_{t,t}=I$.

\paragraph{Assumptions.}
All assumptions are imposed from some iteration $i_0\geq0$ onward, which allows
for early transients.

\begin{assumption}[Contraction of the approximate system]
\label{ass:contr}
There exist $\tilde a\geq1$ and $\tilde\lambda\in\mathbb R$ such that for every
iteration $i\geq i_0$ and all $0\le j<k\le T$,
\[
    \big\|\widetilde A_k^{(i)}\widetilde A_{k-1}^{(i)}\cdots\widetilde A_{j+1}^{(i)}\big\|_2
    \le\tilde a\,e^{\tilde\lambda(k-j)} .
\]
\end{assumption}

This is the predictability condition of \citet{gonzalez2025predictability},
applied to the approximate blocks the solver actually multiplies rather than to
the true Jacobians. Only products of consecutive blocks are needed.

\begin{assumption}[Lipschitz Jacobians]
\label{ass:lip}
There exists $L\geq0$ such that $\|A_t(x)-A_t(y)\|_2\le L\|x-y\|_2$ for all
$t$ and all $x,y\in\mathbb R^D$.
\end{assumption}

\begin{assumption}[Block error along the current trajectory error]
\label{ass:rho}
There exists $\rho\geq0$ such that $\rho_i\le\rho$ for every iteration
$i\geq i_0$, where
\[
    \rho_i:=\frac{\big\|(\widetilde{\mathbf J}^{(i)}-\mathbf J^{(i)})\,e^{(i)}\big\|_\infty}{\|e^{(i)}\|_\infty}
    =\frac{\max_t\big\|(\widetilde A_t^{(i)}-A_t^{(i)})\,e_{t-1}^{(i)}\big\|_2}{\|e^{(i)}\|_\infty},
\]
with $\rho_i:=0$ if $e^{(i)}=0$.
\end{assumption}

The quantity $\rho_i$ measures the block error only along the current trajectory error. It
satisfies $\rho_i\le\sigma_i:=\max_t\|\widetilde A_t^{(i)}-A_t^{(i)}\|_2$, the
spectral-norm block error, and can be much smaller. Since it depends on
$\mathbf x^\star$, it can only be measured a posteriori.

\begin{assumption}[Relative compression error]
\label{ass:comp}
There exists $\eta\geq0$ such that at every iteration $i\geq i_0$ and at every
node $\nu$,
\[
    \|\Delta_\nu\|_2\le\eta\,\tilde a\,e^{\tilde\lambda|\nu|} .
\]
\end{assumption}

By Assumption~\ref{ass:contr}, the right-hand side bounds the size of the exact
product over the segment of $\nu$, so Assumption~\ref{ass:comp} states that each
merge loses at most a fraction $\eta$ of the largest size this product can
reach. It is implied by the stronger condition
$\|\Delta_\nu\|_2\le\eta\,\|\Pi_{k,j}\|_2$.
For truncated-SVD compression with fixed diagonal baseline $\gamma I$,
$\|\Delta_\nu\|_2$ equals the largest discarded singular value of
$X_\nu-\gamma^{|\nu|}I$, so Assumption~\ref{ass:comp} requires this value to be
small. When rank is preserved under multiplication, as in the purely low-rank
case ($\gamma=0$), no compression is needed and $\eta=0$. For randomized
compression, we expect the assumption to hold approximately when the low-rank
parts of the products have rapidly decaying spectra, but we do not prove this.

\paragraph{Effective Lyapunov exponent with compression.}
As in Theorem~\ref{thm:cvg_bound}, define
\[
    \lambda_c:=\tilde\lambda+\log\big(1+\tilde a^2\eta\big)\ \geq\ \tilde\lambda,
\]
with equality when $\eta=0$.

\paragraph{A sufficient condition from the true dynamics.}
Assumptions~\ref{ass:contr} and~\ref{ass:rho} follow from predictability of the
true dynamics together with a bound on the spectral-norm block error.

\begin{proposition}[Predictability and spectral-norm error]
\label{prop:sufficient}
Suppose that for every iteration $i\geq i_0$ and all $0\le j<k\le T$,
$\|A_k^{(i)}\cdots A_{j+1}^{(i)}\|_2\le a\,e^{\lambda(k-j)}$ with $a\geq1$, and
that $\sigma_i\le\sigma$. Then Assumption~\ref{ass:contr} holds with
$\tilde a=a$ and $\tilde\lambda=\log(e^\lambda+a\sigma)$, and
Assumption~\ref{ass:rho} holds with $\rho=\sigma$.
\end{proposition}

\begin{proof}
Write $\widetilde A_\ell=A_\ell+\delta_\ell$ with $\|\delta_\ell\|_2\le\sigma$,
and let $n=k-j$. Expanding $\widetilde A_k\cdots\widetilde A_{j+1}$ gives a sum
over subsets $S\subseteq\{j{+}1,\dots,k\}$ indicating where a factor
$\delta_\ell$ replaces $A_\ell$. For $|S|=m$, the term consists of $m$ factors
$\delta_\ell$ separated by at most $m+1$ products of consecutive exact
Jacobians, with total length $n-m$, so its norm is at most
$a^{m+1}e^{\lambda(n-m)}\sigma^m$, using $a\geq1$. Summing over subsets,
\[
    \big\|\widetilde A_k\cdots\widetilde A_{j+1}\big\|_2
    \le\sum_{m=0}^{n}\binom nm a^{m+1}e^{\lambda(n-m)}\sigma^m
    =a\big(e^\lambda+a\sigma\big)^n .
\]
Finally, $\rho_i\le\sigma_i\le\sigma$.
\end{proof}

This condition is convenient but loose: in our experiments, the blocks are poor
in spectral norm, so $\log(e^\lambda+a\sigma)>0$, while the approximate products
do contract and $\rho_i$ is small (Appendix~\ref{app:conv_diagnostics}).

\subsection{One-step error bound for compressed scans}
\label{app:conv_onestep}

We first show that the compressed scan applies a linear operator to the
residual, and identify its blocks.

\begin{lemma}[Linear representation of the compressed scan]
\label{lem:scan_linear}
At each iteration, $\Delta\mathbf x^{(i+1)}=-\mathbf P\,\mathbf r(\mathbf x^{(i)}_{1:T})$,
where $\mathbf P$ is block lower triangular with blocks $P_{t,s}$ ($s\le t$)
and $P_{t,t}=I$. For $s<t$,
\[
    P_{t,s}=\widehat M_{G_r}\cdots\widehat M_{G_1},
\]
where $G_1,\dots,G_r$ are nodes or leaves whose segments partition
$\{s{+}1,\dots,t\}$ into consecutive blocks, ordered in time. In particular,
$P_{t,t-1}=\widetilde A_t$. Without compression, $P_{t,s}=\Pi_{t,s}$ and
$\mathbf P=\widetilde{\mathbf J}^{-1}$.
\end{lemma}

\begin{proof}
We show by induction over the merge tree that the offset of the summary of any
segment $\{p,\dots,q\}$ is $\sum_{s=p}^{q}Q_s\,c_s$, where $Q_q=I$ and, for
$s<q$, $Q_s$ is a product of stored matrices of nodes or leaves whose segments
partition $\{s{+}1,\dots,q\}$. For a leaf this holds with $Q_q=I$. For a merge
of a left segment $a=\{p,\dots,m\}$ and a right segment
$b=\{m{+}1,\dots,q\}$, the new offset is $\widehat M_b c_a+c_b$. Coefficients of
leaves in $b$ are unchanged, and coefficients of leaves $s\in a$ are multiplied
on the left by $\widehat M_b$. Since the segment of $b$ is
$\{m{+}1,\dots,q\}$ and the inductive coefficient of $s$ in $a$ covers
$\{s{+}1,\dots,m\}$, the product covers $\{s{+}1,\dots,q\}$. Applying this to
the tree of the prefix $\{1,\dots,t\}$ and using
$\Delta x_t^{(i+1)}=c_{1:t}$ with $c_s=-r_s$ gives the claim. For $s=t-1$, the
only partition of $\{t\}$ is the leaf itself, so $P_{t,t-1}=\widetilde A_t$.
Without compression, every stored matrix is the exact product over its segment,
so $P_{t,s}=\Pi_{t,s}$. These are exactly the blocks of
$\widetilde{\mathbf J}^{-1}$ \citep{gonzalez2025unifying}.
\end{proof}

Lemma~\ref{lem:scan_linear} also gives a direct proof of
Theorem~\ref{thm:global_recovery} for any scan tree: if $c_s=0$ for all
$s\le i$, then $\Delta x_{i+1}^{(i+1)}=\sum_{s\le i+1}P_{i+1,s}c_s=c_{i+1}$,
since $P_{i+1,i+1}=I$, regardless of compression.

We measure how far the compressed scan is from the exact inverse through the
compression error $\mathbf E:=\mathbf P-\widetilde{\mathbf J}^{-1}$, with blocks
$\mathbf E_{t,s}=P_{t,s}-\Pi_{t,s}$, and the \emph{consistency defect}
\[
    \Gamma:=I-\mathbf P\,\widetilde{\mathbf J}.
\]

\begin{lemma}[Structure of the consistency defect]
\label{lem:gamma}
$\Gamma$ is strictly block lower triangular, with $\Gamma_{t,t}=\Gamma_{t,t-1}=0$
and, for $s<t$,
\[
    \Gamma_{t,s}=P_{t,s+1}\widetilde A_{s+1}-P_{t,s}.
\]
In particular, $\Gamma=0$ without compression.
\end{lemma}

\begin{proof}
Write $\widetilde{\mathbf J}=I-\widetilde{\mathbf N}$, where
$\widetilde{\mathbf N}$ has blocks $\widetilde A_u$ at positions $(u,u-1)$.
Then $(\mathbf P\widetilde{\mathbf N})_{t,s}=P_{t,s+1}\widetilde A_{s+1}$ for
$s<t$ and $0$ otherwise, so
$(\mathbf P\widetilde{\mathbf J})_{t,s}=P_{t,s}-P_{t,s+1}\widetilde A_{s+1}$ for
$s<t$, and $P_{t,t}=I$ for $s=t$. Subtracting from $I$ gives the formula. For
$s=t-1$, $P_{t,t}\widetilde A_t-P_{t,t-1}=\widetilde A_t-\widetilde A_t=0$ by
Lemma~\ref{lem:scan_linear}. Without compression,
$P_{t,s+1}\widetilde A_{s+1}=\Pi_{t,s+1}\widetilde A_{s+1}=\Pi_{t,s}=P_{t,s}$.
\end{proof}

\begin{proposition}[One-step bound with compressed scan]
\label{prop:onestep}
Under Assumption~\ref{ass:lip}, at every iteration $i$,
\[
    \|e^{(i+1)}\|_\infty
    \le
    \|\Gamma\|_\infty\,\|e^{(i)}\|_\infty
    +\|\mathbf P\|_\infty\Big(\rho_i\,\|e^{(i)}\|_\infty+\tfrac L2\|e^{(i)}\|_\infty^2\Big).
\]
With an exact scan ($\mathbf P=\widetilde{\mathbf J}^{-1}$, $\Gamma=0$) and
$\rho_i$ replaced by its upper bound $\sigma_i$, this is Proposition~3 of
\citet{gonzalez2025unifying}, stated in the per-step norm.
\end{proposition}

\begin{proof}
Drop the superscript $(i)$ and let $e=e^{(i)}$. Since $\mathbf r(\mathbf x^\star)=0$,
for each $t\ge2$,
\[
    r_t(\mathbf x^{(i)})
    =e_t-\big(f_t(x_{t-1}^{(i)})-f_t(x_{t-1}^\star)\big)
    =e_t-A_t\,e_{t-1}+R_t,
\]
where, by Assumption~\ref{ass:lip} and Taylor's theorem,
$R_t:=f_t(x^\star_{t-1})-f_t(x^{(i)}_{t-1})+A_t e_{t-1}$ satisfies
$\|R_t\|_2\le\tfrac L2\|e_{t-1}\|_2^2$. Also $r_1(\mathbf x^{(i)})=e_1$. Hence
$\mathbf r(\mathbf x^{(i)})=\mathbf J e+R$ with $\|R\|_\infty\le\tfrac L2\|e\|_\infty^2$.
The update $e^{(i+1)}=e-\mathbf P\,\mathbf r(\mathbf x^{(i)})$ then gives
\[
    e^{(i+1)}
    =(I-\mathbf P\mathbf J)e-\mathbf PR
    =\Gamma e+\mathbf P\big((\widetilde{\mathbf J}-\mathbf J)e-R\big).
\]
By definition of $\rho_i$,
$\|(\widetilde{\mathbf J}-\mathbf J)e\|_\infty=\rho_i\|e\|_\infty$. Taking norms
concludes.
\end{proof}

The proof applies $\widetilde{\mathbf J}-\mathbf J$ only to the current trajectory error,
which is why the bound involves $\rho_i$ rather than the spectral-norm error
$\sigma_i$.

\subsection{Bounds on compressed products}
\label{app:conv_products}

The following lemma makes part~(i) of Theorem~\ref{thm:cvg_bound} precise:
while products of approximate blocks grow at most at rate $\tilde\lambda$
(Assumption~\ref{ass:contr}), their compressed counterparts used by the scan
grow at most at rate $\lambda_c$.

\begin{lemma}[Compressed products]
\label{lem:comp_products}
Under Assumptions~\ref{ass:contr} and~\ref{ass:comp}, for every iteration
$i\geq i_0$ and all $s<t$, with $n=t-s$,
\begin{align*}
    \|P_{t,s}\|_2
    &\le\tilde a\,e^{\tilde\lambda n}\big(1+\tilde a^2\eta\big)^{n-1}
    \le\tilde a\,e^{\lambda_c n},\\
    \|P_{t,s}-\Pi_{t,s}\|_2
    &\le\tilde a\,e^{\tilde\lambda n}\Big[\big(1+\tilde a^2\eta\big)^{n-1}-1\Big]
    \le\tilde a\big(e^{\lambda_c n}-e^{\tilde\lambda n}\big).
\end{align*}
\end{lemma}

\begin{proof}
\emph{Expansion.} For a node $\nu$, call a set $S$ of nodes in the subtree of
$\nu$ an antichain if no node in $S$ is an ancestor of another. We claim that
\[
    \widehat M_\nu=\sum_{S}\Pi^{(S)}_\nu,
\]
where the sum is over antichains $S$ in the subtree of $\nu$, and
$\Pi^{(S)}_\nu$ is the ordered product over the segment of $\nu$ in which the
segment of each $\mu\in S$ is replaced by the single factor $\Delta_\mu$, and
every remaining time step $u$ contributes $\widetilde A_u$. This holds for
leaves (only $S=\emptyset$). For a node with children $l,r$,
$\widehat M_\nu=\widehat M_r\widehat M_l+\Delta_\nu$. Expanding $\widehat M_r$
and $\widehat M_l$ by induction gives the sum over unions of antichains of the
two subtrees, and $\Delta_\nu$ is the term $S=\{\nu\}$. These are exactly the
antichains of the subtree of $\nu$.

By Lemma~\ref{lem:scan_linear},
$P_{t,s}=\widehat M_{G_r}\cdots\widehat M_{G_1}$, so $P_{t,s}$ expands as a sum
over sets $S$ that are unions of antichains of the subtrees of
$G_1,\dots,G_r$. The term $S=\emptyset$ is $\Pi_{t,s}$.

\emph{Bounding each term.} Fix $S$ with $|S|=j$. The corresponding term
consists of the $j$ factors $\Delta_\mu$, $\mu\in S$, separated by at most
$j+1$ products of consecutive approximate blocks, with segment lengths summing
to $n$. By Assumption~\ref{ass:contr}, each product of consecutive approximate
blocks of length $\ell$ has norm at most $\tilde a\,e^{\tilde\lambda\ell}$
(also for $\ell=0$, since $\tilde a\geq1$), and by
Assumption~\ref{ass:comp}, $\|\Delta_\mu\|_2\le\eta\,\tilde a\,e^{\tilde\lambda|\mu|}$.
Hence the term has norm at most
\[
    \tilde a^{j+1}(\eta\tilde a)^j e^{\tilde\lambda n}
    =\tilde a\,e^{\tilde\lambda n}\big(\tilde a^2\eta\big)^j .
\]

\emph{Counting.} The subtrees of $G_1,\dots,G_r$ contain
$\sum_\ell(|G_\ell|-1)=n-r\le n-1$ nodes in total, so there are at most
$\binom{n-1}{j}$ sets $S$ with $|S|=j$. Summing over $j\ge0$ gives the first
bound, and summing over $j\ge1$ gives the second. The final inequalities use
$e^{\tilde\lambda n}(1+\tilde a^2\eta)^{n-1}\le e^{\tilde\lambda n}(1+\tilde a^2\eta)^{n}=e^{\lambda_c n}$.
\end{proof}

\begin{corollary}[Operator bounds]
\label{cor:operators}
Under Assumptions~\ref{ass:contr} and~\ref{ass:comp}, at every iteration
$i\geq i_0$,
\begin{align*}
    \|\widetilde{\mathbf J}^{-1}\|_\infty&\le\tilde a\,S_T(\tilde\lambda),
    &
    \|\mathbf P\|_\infty&\le\tilde a\,S_T(\lambda_c),\\
    \|\mathbf E\|_\infty&\le\tilde a\big(S_T(\lambda_c)-S_T(\tilde\lambda)\big),
    &
    \|\Gamma\|_\infty&\le 2\tilde a\big(S_T(\lambda_c)-S_T(\tilde\lambda)\big).
\end{align*}
If $\lambda_c<0$, then
\[
    \|\mathbf P\|_\infty\le\frac{\tilde a}{1-e^{\lambda_c}},
    \qquad
    \|\Gamma\|_\infty\le\frac{2\tilde a^3\eta\,e^{\tilde\lambda}}{(1-e^{\tilde\lambda})(1-e^{\lambda_c})} .
\]
\end{corollary}

\begin{proof}
We use the row-sum bound \eqref{eq:app_rowsum}. In row $t$, the block in
column $s$ corresponds to $n=t-s\in\{0,\dots,t-1\}$.

For $\widetilde{\mathbf J}^{-1}$, the blocks are $\Pi_{t,s}$, and
Assumption~\ref{ass:contr} gives
$\sum_{n=0}^{t-1}\tilde a\,e^{\tilde\lambda n}\le\tilde a\,S_T(\tilde\lambda)$,
the diagonal term is $\|I\|_2=1\le\tilde a$. For $\mathbf P$ and $\mathbf E$,
use Lemma~\ref{lem:comp_products} in the same way. The diagonal blocks are $I$
and $0$ respectively.

For $\Gamma$, Lemma~\ref{lem:gamma} gives $\Gamma_{t,s}=0$ for $n\le1$. For
$n\ge2$, since $\Pi_{t,s+1}\widetilde A_{s+1}=\Pi_{t,s}$,
\[
    \Gamma_{t,s}
    =\big(P_{t,s+1}-\Pi_{t,s+1}\big)\widetilde A_{s+1}
    -\big(P_{t,s}-\Pi_{t,s}\big).
\]
The second term is bounded by Lemma~\ref{lem:comp_products}. For the first,
expand $P_{t,s+1}-\Pi_{t,s+1}$ as in the proof of
Lemma~\ref{lem:comp_products}: each term has $j\ge1$ factors $\Delta_\mu$ over
$\{s{+}2,\dots,t\}$, and right-multiplying by $\widetilde A_{s+1}$ only extends
its last product of consecutive approximate blocks. Each term is therefore
bounded by $\tilde a\,e^{\tilde\lambda n}(\tilde a^2\eta)^j$, and there are at
most $\binom{n-2}{j}$ of them, so this term is also bounded by
$\tilde a(e^{\lambda_c n}-e^{\tilde\lambda n})$. Hence
$\|\Gamma_{t,s}\|_2\le2\tilde a(e^{\lambda_c n}-e^{\tilde\lambda n})$, and summing
over $n$ gives the bound.

If $\lambda_c<0$, then $\tilde\lambda<0$ as well. Since each term
$e^{\lambda_c n}-e^{\tilde\lambda n}$ is nonnegative,
\[
    S_T(\lambda_c)-S_T(\tilde\lambda)
    \le\frac{1}{1-e^{\lambda_c}}-\frac{1}{1-e^{\tilde\lambda}}
    =\frac{e^{\lambda_c}-e^{\tilde\lambda}}{(1-e^{\lambda_c})(1-e^{\tilde\lambda})}
    =\frac{\tilde a^2\eta\,e^{\tilde\lambda}}{(1-e^{\lambda_c})(1-e^{\tilde\lambda})},
\]
using $e^{\lambda_c}=e^{\tilde\lambda}(1+\tilde a^2\eta)$.
\end{proof}

\subsection{Proof of Theorem~\ref{thm:cvg_bound}}
\label{app:conv_proof}

We prove a general version that does not assume $\lambda_c<0$.
Theorem~\ref{thm:cvg_bound} is the special case $\lambda_c<0$.

\begin{theorem}[General form of Theorem~\ref{thm:cvg_bound}]
\label{thm:cvg_general}
Let Assumptions~\ref{ass:contr}--\ref{ass:comp} hold from iteration $i_0$
onward, and define
\[
    c_0^{(T)}:=\tilde a\,\rho\,S_T(\lambda_c)+2\tilde a\big(S_T(\lambda_c)-S_T(\tilde\lambda)\big),
    \qquad
    c_1^{(T)}:=\tfrac{\tilde aL}{2}\,S_T(\lambda_c).
\]
Then:
\begin{enumerate}
    \item[(i)] For every iteration $i\geq i_0$ and all $s<t$,
    $\|P_{t,s}\|_2\le\tilde a\,e^{\lambda_c(t-s)}$.
    \item[(ii)] For every iteration $i\geq i_0$,
    $\|e^{(i+1)}\|_\infty\le\big(c_0^{(T)}+c_1^{(T)}\|e^{(i)}\|_\infty\big)\|e^{(i)}\|_\infty$.
    \item[(iii)] If $c_0^{(T)}<1$ and
    $\|e^{(i_0)}\|_\infty<(1-c_0^{(T)})/c_1^{(T)}$, then
    $\|e^{(i)}\|_\infty\le q^{\,i-i_0}\|e^{(i_0)}\|_\infty$ for all $i\geq i_0$,
    with $q:=c_0^{(T)}+c_1^{(T)}\|e^{(i_0)}\|_\infty<1$.
\end{enumerate}
If $\lambda_c<0$, then $c_0^{(T)}\le c_0$ and $c_1^{(T)}\le c_1$, with $c_0$
and $c_1$ as in Theorem~\ref{thm:cvg_bound}, which are independent of $T$.
\end{theorem}

\begin{proof}
\emph{(i)} This is Lemma~\ref{lem:comp_products}.

\emph{(ii)} Insert the bounds of Corollary~\ref{cor:operators} into
Proposition~\ref{prop:onestep}, and use $\rho_i\le\rho$:
\[
    \|e^{(i+1)}\|_\infty
    \le\Big(2\tilde a\big(S_T(\lambda_c)-S_T(\tilde\lambda)\big)
    +\tilde a\,S_T(\lambda_c)\,\rho
    +\tilde a\,S_T(\lambda_c)\tfrac L2\|e^{(i)}\|_\infty\Big)\|e^{(i)}\|_\infty .
\]

\emph{(iii)} We show by induction on $i\geq i_0$ that
$\|e^{(i)}\|_\infty\le\|e^{(i_0)}\|_\infty$ and
$\|e^{(i)}\|_\infty\le q^{\,i-i_0}\|e^{(i_0)}\|_\infty$. Both hold for $i=i_0$.
If they hold for $i$, then by (ii),
\[
    \|e^{(i+1)}\|_\infty
    \le\big(c_0^{(T)}+c_1^{(T)}\|e^{(i_0)}\|_\infty\big)\|e^{(i)}\|_\infty
    =q\,\|e^{(i)}\|_\infty
    \le q^{\,i+1-i_0}\|e^{(i_0)}\|_\infty ,
\]
and since $q<1$, also $\|e^{(i+1)}\|_\infty\le\|e^{(i_0)}\|_\infty$. The
condition $q<1$ is equivalent to $\|e^{(i_0)}\|_\infty<(1-c_0^{(T)})/c_1^{(T)}$.

Finally, if $\lambda_c<0$, Corollary~\ref{cor:operators} gives
\[
    c_0^{(T)}\le\frac{\tilde a\,\rho}{1-e^{\lambda_c}}
    +\frac{2\tilde a^3\eta\,e^{\tilde\lambda}}{(1-e^{\tilde\lambda})(1-e^{\lambda_c})}=c_0,
    \qquad
    c_1^{(T)}\le\frac{\tilde aL}{2(1-e^{\lambda_c})}=c_1 .
\]
Since (ii) and (iii) remain valid when $c_0^{(T)},c_1^{(T)}$ are replaced by
the larger constants $c_0,c_1$, this proves Theorem~\ref{thm:cvg_bound}.
\end{proof}

\subsection{Consequences and limitations}
\label{app:conv_consequences}

\paragraph{Special cases.}
\begin{itemize}
    \item \emph{Exact Newton, exact scan} ($\rho=\eta=0$): $c_0=0$, and (ii)
    reads $\|e^{(i+1)}\|_\infty\le c_1\|e^{(i)}\|_\infty^2$, i.e.\ local
    quadratic convergence.
    \item \emph{Approximate blocks, exact scan} ($\eta=0$): $\Gamma=0$,
    $\lambda_c=\tilde\lambda$, and $c_0=\tilde a\rho/(1-e^{\tilde\lambda})$.
    Compared with Proposition~3 of \citet{gonzalez2025unifying}, the inverse
    is bounded independently of $T$,
    $\|\widetilde{\mathbf J}^{-1}\|_\infty\le\tilde a/(1-e^{\tilde\lambda})$,
    even when individual blocks satisfy $\|\widetilde A_t\|_2>1$, and the block
    error enters only along the current trajectory error. This case includes diagonal
    quasi-Newton.
    \item \emph{Exact blocks, compressed scan} ($\rho=0$, $\eta>0$):
    $c_0=\frac{2\tilde a^3\eta e^{\tilde\lambda}}{(1-e^{\tilde\lambda})(1-e^{\lambda_c})}>0$,
    so the bound guarantees only a linear rate, $O(\eta)$.
    \item \emph{Vanishing block error}: if $\eta=0$ and $\rho_i\to0$,
    Proposition~\ref{prop:onestep} gives a per-iteration factor
    $\|\mathbf P\|_\infty(\rho_i+\tfrac L2\|e^{(i)}\|_\infty)\to0$, i.e.\
    superlinear convergence.
\end{itemize}

\paragraph{Dependence on $T$.}
If $\lambda_c<0$, all constants are independent of $T$. If $\lambda_c=0$,
then $S_T(\lambda_c)=T$, so $c_0^{(T)}$ and $c_1^{(T)}$ grow linearly with $T$.
If $\lambda_c>0$, they grow like $e^{\lambda_c T}$. In both cases,
$c_0^{(T)}<1$ requires $\rho$ and $\eta$ to shrink with $T$, and the region
$\|e^{(i_0)}\|_\infty<(1-c_0^{(T)})/c_1^{(T)}$ shrinks with $T$, so the
guarantee becomes uninformative for long sequences. For exact Newton on a
chaotic system, this recovers the picture of
\citet{gonzalez2025predictability}: quadratic convergence holds, but only in a
region whose size decays exponentially with $T$.

\paragraph{Iteration complexity.}
Under the conditions of (iii), $\|e^{(i)}\|_\infty\le\varepsilon$ as soon as
\[
    i-i_0\ \ge\ \frac{\log\big(\|e^{(i_0)}\|_\infty/\varepsilon\big)}{\log(1/q)} .
\]
When $\lambda_c<0$ and the states are bounded, this is independent of $T$. Since
each iteration is a parallel scan of depth $O(\log T)$, this phase costs
$O(\log T)$ parallel depth. If accuracy is instead measured in the Euclidean
norm over the whole trajectory, $\|e\|_2\le\sqrt T\,\|e\|_\infty$ requires
$\|e\|_\infty\le\varepsilon/\sqrt T$, which adds $\tfrac12\log T/\log(1/q)$
iterations and gives $O(\log^2 T)$ depth, as in
\citet{gonzalez2025predictability}.

\paragraph{Locality and damped steps.}
The result requires $\|e^{(i_0)}\|_\infty<(1-c_0)/c_1$. We do not bound the
number of iterations needed to reach such an iteration from an arbitrary
initialization. Theorem~\ref{thm:global_recovery} only guarantees that the
error vanishes after at most $T$ iterations. When $L=0$ (linear dynamics),
$c_1=0$ and the condition reduces to $c_0<1$. The analysis covers full steps.
For a step scaled by $\alpha\in(0,1]$,
$e^{(i+1)}=(1-\alpha)e^{(i)}+\alpha\big(e^{(i)}-\mathbf P\mathbf r\big)$, so
the per-iteration factor becomes $1-\alpha+\alpha(c_0+c_1\|e^{(i)}\|_\infty)$,
which still contracts under the same conditions, more slowly. Iterations where
a line search replaces the quasi-Newton step altogether are not covered.

\paragraph{Limitations.}
Theorem~\ref{thm:cvg_bound} applies to any approximate blocks satisfying
Assumptions~\ref{ass:contr}--\ref{ass:comp}, which are imposed along the
iterates. In particular, it does not show that the Broyden secant updates keep
$\rho$ small enough for $c_0<1$. It describes how convergence depends on the
block approximation error, whatever its source. The quantity $\rho_i$ depends on
$\mathbf x^\star$ and can only be measured a posteriori. The constants are worst-case and, in our
experiments, far from tight (Appendix~\ref{app:conv_diagnostics}). The main
slack comes from the row-sum bound \eqref{eq:app_rowsum}, which adds the
contributions of all upstream time steps as if they were aligned, and from the
powers of $\tilde a$, which come from bounding each product of consecutive
blocks separately. Finally, Assumption~\ref{ass:comp} is stated per merge. We do
not relate $\eta$ to the compression rank $m$ or to the spectrum of the dynamics
Jacobians.

\section{Empirical diagnostics of the convergence analysis}
\label{app:conv_diagnostics}

Theorem~\ref{thm:cvg_bound} measures block quality by the error along the
current trajectory error, $\rho$, rather than by the spectral-norm error $\sigma$ used in
prior analyses \citep{gonzalez2025unifying}. Here we measure both, together with
the other quantities in the theorem, along the iterates of the solvers. The
main finding is that $\sigma$ does not track convergence while $\rho$ does.

\subsection{Setup}

We use the low-rank RNNs of Appendix~\ref{app:low_rank_rnn} (standard and
MLP variants, $D=200$, $\Delta t\in\{0.5,3\}$) and the hierarchical Gibbs
sampler of Appendix~\ref{app:gibbs} ($S\in\{10,50,100\}$, $D=2S+2$), with
$T=10^3$ and $8$ chains per method and setting. On the RNNs, we compare
Accelerated Parallel Broyden, Parallel Broyden, and diagonal quasi-Newton. On
the Gibbs sampler, we compare the two Broyden variants for $S\in\{10,50\}$, and
run Accelerated Parallel Broyden alone for $S=100$, with a single chain. All
other settings (damping, rank, diagonal initialization, line search) are those
of the main experiments.

At each iteration, we rebuild the blocks $\widetilde A_t^{(i)}$ offline from
the solver's state. The rebuilt blocks reproduce the solver's steps to a median
relative error below $4\cdot10^{-3}$. Iterations where the backtracking line
search of Accelerated Parallel Broyden replaced the quasi-Newton step are
excluded. Errors $e^{(i)}$ are measured against the
sequential trajectory. We record:
\begin{itemize}
    \item the spectral-norm block error $\sigma_i=\max_t\|\widetilde A_t^{(i)}-A_t^{(i)}\|_2$,
    \item the error-direction quantity $\rho_i$ of Assumption~\ref{ass:rho},
    \item the compression error at every merge of the scan tree,
    \item the per-iteration error ratio $\|e^{(i+1)}\|_\infty/\|e^{(i)}\|_\infty$.
\end{itemize}
We estimate $\tilde a$ and $\tilde\lambda$ by fitting
Assumption~\ref{ass:contr} to the products over the segments of the merge tree,
and compute $\eta$ and the $\rho$ term of $c_0$ at the replayed iterations. To
isolate the effect of damping, we also evaluate $\sigma_i$ and $\rho_i$ with the
undamped blocks along the same iterates.

\subsection{$\sigma$ does not track convergence, $\rho$ does}

Table~\ref{tab:diag_rho} reports both measures of block quality at the last
iteration.

\begin{table}[h]
\centering
\caption{Spectral-norm block error $\sigma_i$ and error-direction quantity
$\rho_i$ at the last iteration, and median iteration counts over the $8$ diagnostic chains. Diagonal quasi-Newton was run
on the RNNs only.}
\label{tab:diag_rho}
\small
\begin{tabular}{llccc}
\toprule
Setting & Method & $\sigma_i$ & $\rho_i$ & Iterations \\
\midrule
Std RNN, $\Delta t=0.5$ & Accel.    & $0.885$ & $0.025$ & $9$ \\
                        & Broyden   & $0.885$ & $0.057$ & $13$ \\
                        & Diag.\ QN & $0.891$ & $0.118$ & $37.5$ \\
\midrule
MLP RNN, $\Delta t=0.5$ & Accel.    & $1.074$ & $0.041$ & $9.5$ \\
                        & Broyden   & $0.991$ & $0.044$ & $12$ \\
                        & Diag.\ QN & $0.996$ & $0.131$ & $35$ \\
\midrule
Std RNN, $\Delta t=3$   & Accel.    & $5.32$ & $0.076$ & $9.5$ \\
                        & Broyden   & $5.32$ & $0.333$ & $34^\dagger$ \\
                        & Diag.\ QN & $5.38$ & $0.734$ & $50.5$ \\
\midrule
MLP RNN, $\Delta t=3$   & Accel.    & $6.62$ & $0.085$ & $10.5$ \\
                        & Broyden   & $6.62$ & $0.320$ & $38$ \\
                        & Diag.\ QN & $6.66$ & $0.716$ & $48.5$ \\
\midrule
Gibbs, $S=10$           & Accel.    & $129.6$ & $0.337$ & $23$ \\
                        & Broyden   & $129.6$ & $0.907$ & $58.5$ \\
\midrule
Gibbs, $S=50$           & Accel.    & $73.5$ & $0.172$ & $27.5$ \\
                        & Broyden   & $73.5$ & $0.766$ & $87.5$ \\
\midrule
Gibbs, $S=100$          & Accel.    & $71.7$ & $0.097$ & $31$ \\
\bottomrule
\end{tabular}

\smallskip
{\footnotesize $^\dagger$\,$7$ of $8$ chains converged. All other settings
$8/8$.}
\end{table}

\paragraph{The spectral-norm error is large and the same for all methods.}
At the last iteration, $\sigma_i$ is about $0.9$--$1.1$ for the RNNs at
$\Delta t=0.5$, $5.3$--$6.6$ at $\Delta t=3$, and $72$--$130$ for the Gibbs
sampler. Relative to the true block, the error
$\max_t\|\widetilde A_t-A_t\|_2/\|A_t\|_2$ is about $0.54$ at $\Delta t=0.5$,
$0.98$ at $\Delta t=3$, and essentially $1$ for the Gibbs sampler. Since
$\sigma\geq0.88$ in every setting, $\log(e^\lambda+a\sigma)>0$, and the
sufficient condition of Proposition~\ref{prop:sufficient} fails everywhere.
More importantly, $\sigma_i$ does not distinguish the methods: in every
setting, it is nearly identical for Accelerated Parallel Broyden, Parallel
Broyden, and diagonal quasi-Newton (within $9\%$, and within $1\%$ outside the
MLP RNN at $\Delta t=0.5$), although their iteration counts differ by up to
$5.3\times$. This is expected: the worst-case error is set by the part of the
off-diagonal coupling that each approximation misses in some direction.
Diagonal quasi-Newton discards this coupling entirely, and rank-$m$ secant
information only corrects it along the directions the iterates have explored.

\paragraph{The block error along the current trajectory error is small and tracks
convergence.}
In contrast, $\rho_i$ is $7$--$740$ times smaller than $\sigma_i$. At the last
iteration, the ordering of $\rho_i$ matches the ordering of iteration counts in
every setting: Accelerated Parallel Broyden has the smallest $\rho_i$ and the
fewest iterations, followed by Parallel Broyden and then diagonal quasi-Newton.
The size of the gap also tracks convergence: the ratio of $\rho_i$ between the
two Broyden variants is $1.1$ for the MLP RNN at $\Delta t=0.5$, where their
iteration counts are close ($9.5$ versus $12$), and $3.7$--$4.5$ where
Accelerated Parallel Broyden is $3$--$3.6$ times faster.

\paragraph{Along the solve.}
Since the methods take different numbers of iterations, we also compare them on
the RNNs at the same stage of the solve, by grouping iterations according to
their root-mean-square residual (Table~\ref{tab:diag_along}). Two conclusions
hold at every stage. First, $\sigma_i$ is the same for all three methods
throughout: its relative value stays within about $0.02$ across methods, at
$0.53$--$0.62$ for $\Delta t=0.5$ and $0.97$--$0.99$ for $\Delta t=3$. Second,
diagonal quasi-Newton has the largest $\rho_i$ at every stage, by a wide margin
($0.13$--$0.17$ versus $0.05$--$0.10$ for the Broyden variants at
$\Delta t=0.5$, and $0.79$--$0.91$ versus $0.09$--$0.49$ at $\Delta t=3$). The
ordering between the two Broyden variants, however, emerges only near
convergence. Early in the solve, the $\rho_i$ of Accelerated Parallel Broyden
is as large as that of Parallel Broyden or larger, and it drops clearly below
only in the last decade of residual before the tolerance. This is consistent
with the local nature of Theorem~\ref{thm:cvg_bound}. It also means that $\rho_i$
does not explain the advantage of Accelerated Parallel Broyden over the whole
solve: part of this advantage comes from how quickly it reaches the regime of
small $\rho_i$. Accelerated Parallel Broyden also shows transient spikes, with
$\rho_i>1$ in $2$--$18\%$ of iterations, which is why the assumptions are imposed
only from some iteration $i_0$ onward.

\begin{table}[h]
\centering
\caption{Median $\rho_i$ for the two Broyden variants on the RNNs at
$\Delta t=3$, grouped by the root-mean-square residual
$\|\mathbf r\|_2/\sqrt{TD}$. Ranges are over the
standard and MLP models.}
\label{tab:diag_along}
\small
\begin{tabular}{lcccc}
\toprule
RMS residual & $\geq10^{-1}$ & $[10^{-2},10^{-1})$ & $[10^{-3},10^{-2})$ & $[10^{-4},10^{-3})$ \\
\midrule
Parallel Broyden             & $0.46$        & $0.30$        & $0.24$--$0.27$ & $0.23$--$0.27$ \\
Accelerated Parallel Broyden & $0.48$--$0.49$ & $0.36$--$0.40$ & $0.19$--$0.29$ & $0.09$--$0.11$ \\
\bottomrule
\end{tabular}
\end{table}

\subsection{Damping reduces $\rho$, not $\sigma$}

For Parallel Broyden, damping leaves $\sigma_i$ essentially unchanged but lowers
$\rho_i$ in every setting, and nearly eliminates iterations with $\rho_i>1$
(Table~\ref{tab:diag_damping}). Damping therefore improves the blocks where it
matters, which is consistent with its effect on robustness
(Appendix~\ref{app:broyden_damping}). For Accelerated Parallel Broyden, damping
has little effect on $\rho_i$, and the undamped value is often slightly smaller.
For diagonal quasi-Newton, damping again leaves $\sigma_i$ unchanged ($6.66$
versus $6.61$ for the MLP RNN at $\Delta t=3$) but increases $\rho_i$ ($0.72$
versus $0.22$), as expected: it shrinks the blocks for stability rather than
making them more accurate.

\begin{table}[h]
\centering
\caption{Effect of damping on Parallel Broyden: $\sigma_i$ and $\rho_i$ at the
last iteration and fraction of iterations with $\rho_i>1$, with damped and
undamped blocks along the same iterates.}
\label{tab:diag_damping}
\small
\begin{tabular}{lcccccc}
\toprule
 & \multicolumn{2}{c}{$\sigma_i$ (last)} & \multicolumn{2}{c}{$\rho_i$ (last)} & \multicolumn{2}{c}{Frac.\ $\rho_i>1$} \\
\cmidrule(lr){2-3}\cmidrule(lr){4-5}\cmidrule(lr){6-7}
Setting & damped & undamped & damped & undamped & damped & undamped \\
\midrule
Std RNN, $\Delta t=0.5$ & $0.885$ & $0.885$ & $0.057$ & $0.066$ & $0$    & $0$ \\
MLP RNN, $\Delta t=0.5$ & $0.991$ & $0.990$ & $0.044$ & $0.058$ & $0$    & $0$ \\
Std RNN, $\Delta t=3$   & $5.32$  & $5.56$  & $0.333$ & $0.488$ & $0$    & $0.095$ \\
MLP RNN, $\Delta t=3$   & $6.62$  & $6.62$  & $0.320$ & $0.537$ & $0$    & $0.034$ \\
Gibbs, $S=10$           & $129.6$ & $129.6$ & $0.907$ & $1.279$ & $0.20$ & $0.54$ \\
Gibbs, $S=50$           & $73.5$  & $73.5$  & $0.766$ & $0.885$ & $0.05$ & $0.31$ \\
\bottomrule
\end{tabular}
\end{table}

\subsection{Other quantities in the theorem}

\paragraph{The approximate system contracts.}
The fitted rates satisfy $\tilde\lambda<0$ in every setting:
$\tilde\lambda\in[-0.031,-0.016]$ for the RNNs at $\Delta t=0.5$,
$[-0.134,-0.038]$ at $\Delta t=3$, and $[-0.6,-0.12]$ for the Gibbs sampler.
Assumption~\ref{ass:contr} therefore holds even though its sufficient condition
in terms of $\sigma$ fails.

\paragraph{Compression errors are small.}
For the RNNs, with the rank-revealing randomized compression of
Appendix~\ref{app:rnn}, $\eta$ ranges from about $10^{-4}$ to $0.11$ across
iterations. For the Gibbs sampler, the approximation is purely low rank, so no
compression is required (Appendix~\ref{app:gibbs}) and $\eta=0$ in exact
arithmetic. The measured values, between $4\cdot10^{-8}$ and $4\cdot10^{-4}$,
reflect floating-point error.
Measuring the compression error relative to the exact product
$\|\Pi_{k,j}\|_2$ instead would give spuriously large values on long segments
whose exact product is numerically zero, which is why
Assumption~\ref{ass:comp} uses the bound $\tilde a e^{\tilde\lambda|\nu|}$.
Consistent with Appendix~\ref{app:compression_impact}, the choice of
compression scheme does not affect convergence.

\subsection{The constants are not tight}

Table~\ref{tab:diag_c0} compares the bound with the observed contraction. The
$\rho$ term of $c_0$ alone exceeds $1$ in every setting, so the theorem does not
certify the observed convergence. In addition, $\lambda_c\geq0$ at a substantial
fraction of iterations for the RNNs, because the compression penalty
$\log(1+\tilde a^2\eta)$ is comparable to $|\tilde\lambda|$ at $\Delta t=0.5$
and the fitted $\tilde a$ is large in early iterations of Accelerated Parallel
Broyden. Convergence is also not monotone: the per-step error increases in
$2$--$33\%$ of iterations. Nevertheless, near convergence the error contracts by
a factor of $0.30$--$0.95$ per iteration, with the same ordering across methods
as $\rho_i$. The main slack is the row-sum bound \eqref{eq:app_rowsum}, which
treats contributions from all upstream time steps as aligned. Since $\tilde a$
and $\tilde\lambda$ are fitted only on segments aligned with the merge tree,
the true constants can only be larger, and the conclusion $c_0>1$ is
unaffected. A bound in terms of $\sigma$ instead of $\rho$ would be larger by
the ratio $\sigma_i/\rho_i$, which is $16$--$740$ for the Broyden variants.

\begin{table}[h]
\centering
\caption{The $\rho$ term of $c_0$ (range over iterations with $\lambda_c<0$),
fraction of iterations with $\lambda_c<0$, observed contraction
$\|e^{(i+1)}\|_\infty/\|e^{(i)}\|_\infty$ (median over the last three
iterations), and fraction of iterations where the error increases.}
\label{tab:diag_c0}
\small
\begin{tabular}{llcccc}
\toprule
Setting & Method & $\rho$ term of $c_0$ & Frac.\ $\lambda_c<0$ & Observed & Frac.\ increase \\
\midrule
Std RNN, $\Delta t=0.5$ & Accel.  & $10.7$--$45.3$ & $0.83$ & $0.35$ & $0.18$ \\
                        & Broyden & $2.1$--$10.9$  & $0.50$ & $0.57$ & $0.25$ \\
MLP RNN, $\Delta t=0.5$ & Accel.  & $1.9$--$7.1$   & $0.67$ & $0.41$ & $0.18$ \\
                        & Broyden & $6.1$--$20.9$  & $0.50$ & $0.61$ & $0.17$ \\
Std RNN, $\Delta t=3$   & Accel.  & $1.4$--$3.3$   & $0.33$ & $0.30$ & $0.30$ \\
                        & Broyden & $4.1$--$10.8$  & $1.00$ & $0.72$ & $0.33$ \\
MLP RNN, $\Delta t=3$   & Accel.  & $1.7$--$4.6$   & $0.33$ & $0.39$ & $0.28$ \\
                        & Broyden & $3.5$--$7.0$   & $1.00$ & $0.69$ & $0.32$ \\
Gibbs, $S=10$           & Accel.  & $2.5$--$35.3$  & $1.00$ & $0.46$ & $0.11$ \\
                        & Broyden & $1.6$--$124$   & $1.00$ & $0.81$ & $0.22$ \\
Gibbs, $S=50$           & Accel.  & $1.5$--$6.8$   & $1.00$ & $0.54$ & $0.02$ \\
                        & Broyden & $1.5$--$74.8$  & $1.00$ & $0.95$ & $0.07$ \\
Gibbs, $S=100$          & Accel.  & $1.2$--$6.9$   & $1.00$ & $0.69$ & $0.03$ \\
\bottomrule
\end{tabular}
\end{table}

\subsection{Summary}

The spectral-norm error $\sigma$ is not the right measure of block quality: it
is large, it is essentially the same for all three methods although their
iteration counts differ by up to $5.3\times$, and damping leaves it
unchanged although damping helps Parallel Broyden converge. The error along the current
error, $\rho$, is the quantity that tracks convergence: it is $7$--$740$ times
smaller than $\sigma$, diagonal quasi-Newton has the largest $\rho$ at every
stage of the solve, near convergence $\rho$ is ordered as Accelerated Parallel
Broyden $<$ Parallel Broyden $<$ diagonal quasi-Newton in every setting,
matching their iteration counts, and damping reduces it. The approximate system contracts and
compression errors are small. The theorem identifies this mechanism, but its
constants are too conservative to certify the observed rates.

\section{Broyden updates}
\label{app:broyden}

This section gives the explicit blockwise Broyden updates used to construct the
local dynamics Jacobian approximations in Section~\ref{sec:broyden}. We use the
same notation as in the main text. After computing the iterate
$\mathbf{x}^{(i+1)}_{1:T}$, for each time index $t$, define the trajectory
secant
\[
    p_t^{(i)}
    :=
    x_{t-1}^{(i+1)} - x_{t-1}^{(i)}
\]
and the corresponding finite difference through the local dynamics map
\[
    q_t^{(i)}
    :=
    f_t(x_{t-1}^{(i+1)}) - f_t(x_{t-1}^{(i)}).
\]
The goal is to update an approximation
\[
    \widetilde A_t^{(i)}
    \approx
    A_t^{(i)}
    :=
    \frac{\partial f_t}{\partial x_{t-1}}
    \left(x_{t-1}^{(i)}\right)
\]
using only the secant pair $(p_t^{(i)},q_t^{(i)})$.

\subsection{Standard rank-one Broyden update}
\label{app:broyden_standard}

The standard Broyden update enforces the secant condition
\begin{equation}
    \widetilde A_t^{(i+1)} p_t^{(i)} = q_t^{(i)}.
    \label{eq:app_broyden_secant}
\end{equation}
Since this condition does not uniquely determine the full matrix
$\widetilde A_t^{(i+1)}$, Broyden's method chooses the update that is closest to
the previous approximation in Frobenius norm:
\begin{equation}
    \widetilde A_t^{(i+1)}
    =
    \arg\min_A
    \|A-\widetilde A_t^{(i)}\|_F
    \quad
    \text{s.t.}
    \quad
    Ap_t^{(i)} = q_t^{(i)}.
    \label{eq:app_broyden_min}
\end{equation}
Writing the update as $B=\widetilde A_t^{(i)} + \Delta_t^{(i)}$, the constraint
becomes
\[
    \Delta_t^{(i)}p_t^{(i)}
    =
    q_t^{(i)} - \widetilde A_t^{(i)}p_t^{(i)}.
\]
Let
\[
    y_t^{(i)}
    :=
    q_t^{(i)} - \widetilde A_t^{(i)}p_t^{(i)}
\]
denote the current secant residual. Then the least-change problem is
\[
    \min_{\Delta}
    \|\Delta\|_F
    \quad
    \text{s.t.}
    \quad
    \Delta p_t^{(i)} = y_t^{(i)}.
\]
The minimum-norm solution is rank one:
\[
    \Delta_t^{(i)}
    =
    \frac{
        y_t^{(i)} {p_t^{(i)}}^\top
    }{
        {p_t^{(i)}}^\top p_t^{(i)}
    }.
\]
Thus the blockwise Broyden update is
\begin{equation}
    \boxed{
    \widetilde A_t^{(i+1)}
    =
    \widetilde A_t^{(i)}
    +
    \frac{
        \left(q_t^{(i)}-\widetilde A_t^{(i)}p_t^{(i)}\right)
        {p_t^{(i)}}^\top
    }{
        {p_t^{(i)}}^\top p_t^{(i)}
    }}
    \label{eq:app_broyden_update}
\end{equation}

Starting from a diagonal initialization
\[
    \widetilde A_t^{(0)} = D_t^{(0)},
    \qquad
    D_t^{(0)}=\operatorname{diag}(d_t^{(0)}),
\]
each update in \eqref{eq:app_broyden_update} adds one rank-one correction. After
$i$ updates, the approximation can therefore be written as
\begin{equation}
    \widetilde A_t^{(i)}
    =
    D_t^{(0)}
    +
    U_t^{(i)} {V_t^{(i)}}^\top,
    \label{eq:app_broyden_dplr}
\end{equation}
where the columns of $U_t^{(i)}$ and $V_t^{(i)}$ are the accumulated rank-one
factors. In particular, if
\[
    u_t^{(i)}
    =
    q_t^{(i)}-\widetilde A_t^{(i)}p_t^{(i)},
    \qquad
    v_t^{(i)}
    =
    \frac{p_t^{(i)}}{{p_t^{(i)}}^\top p_t^{(i)}},
\]
then the update adds $u_t^{(i)} {v_t^{(i)}}^\top$ to the current approximation.

\subsection{Limited-memory Broyden updates}
\label{app:broyden_limited_memory}

\paragraph{Simple truncation.}
The standard rank-one Broyden update adds one new low-rank correction at each
outer iteration. To keep the representation size fixed, we use a
limited-memory variant with maximum rank $m$. Throughout this section, we assume
the diagonal baseline is initialized as $\gamma I_D$, so each approximate block
has the form
\[
    \widetilde A_t^{(i)}
    =
    \gamma I_D
    +
    U_t^{(i)} {V_t^{(i)}}^\top,
    \qquad
    U_t^{(i)},V_t^{(i)}\in\mathbb{R}^{D\times m}.
\]
When a new rank-one correction $u_t^{(i)}{v_t^{(i)}}^\top$ is computed and the
memory is already full, we discard the oldest stored correction and append the
newest one:
\[
    U_t^{(i+1)}
    =
    \bigl[
        U_t^{(i)}[:,2],\ldots,U_t^{(i)}[:,m],u_t^{(i)}
    \bigr],
    \qquad
    V_t^{(i+1)}
    =
    \bigl[
        V_t^{(i)}[:,2],\ldots,V_t^{(i)}[:,m],v_t^{(i)}
    \bigr].
\]
Thus the rank remains at most $m$, and applying each block costs
$\mathcal{O}(Dm)$. This simple truncation strategy is the default
limited-memory scheme used in our experiments.

\begin{figure}
    \centering
    \includegraphics[width=\linewidth]{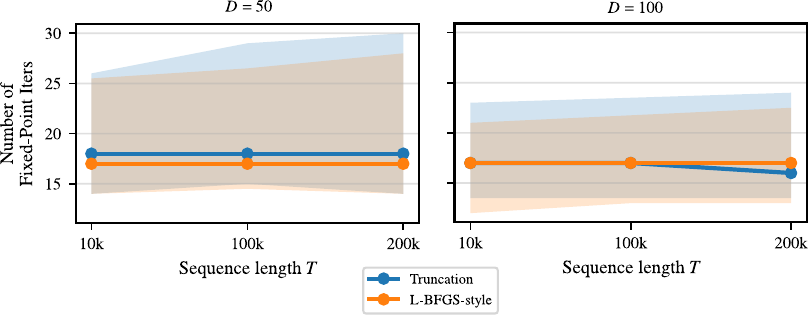}
    \caption{\textbf{Effect of the limited-memory update rule on the low-rank RNN benchmark.}
We compare simple truncation, which discards the oldest rank-one correction once
the rank budget is full, with the L-BFGS-style \texttt{identity\_rescaled}
variant, which rescales the identity baseline from the most recent secant pair
and replays the stored updates. Points denote medians over $32$ random seeds,
and error bars denote interquartile ranges. The rescaled variant does not
substantially improve median performance over simple truncation, but appears
slightly more stable, with slightly smaller upper quartiles.}
    \label{fig:limited}
\end{figure}

\paragraph{L-BFGS-style identity rescaling.}
We also considered an L-BFGS-style scalar-rescaling variant for the identity
baseline. The motivation is that, after truncation, the stored rank-one
corrections contain only the most recent secant information, so the identity
baseline should reflect the scale of the local Jacobian not captured by this
limited memory. For each transition $f_t$, we store the last $m$ secant pairs
$\{(p_{t,j},q_{t,j})\}_{j=i-m}^{i-1}$. At iteration $i$, we choose a scalar
$\gamma_t^{(i)}$ from the most recent pair by fitting the identity baseline to
the secant target:
\[
    \gamma_t^{(i)}
    =
    \arg\min_{\gamma}
    \left\|
        \gamma p_{t,i-1} - q_{t,i-1}
    \right\|_2^2
    =
    \frac{
        \langle p_{t,i-1}, q_{t,i-1}\rangle
    }{
        \|p_{t,i-1}\|_2^2+\varepsilon
    },
\]
with a small $\varepsilon>0$ for numerical stability. We then rebuild the
limited-memory approximation from the rescaled baseline
$\widetilde A_{t,0}^{(i)}=\gamma_t^{(i)}I_D$ by replaying the stored Broyden
updates in chronological order:
\[
    \widetilde A_{t,j+1}^{(i)}
    =
    \widetilde A_{t,j}^{(i)}
    +
    \frac{
        \left(q_{t,j}-\widetilde A_{t,j}^{(i)}p_{t,j}\right)p_{t,j}^\top
    }{
        \|p_{t,j}\|_2^2+\varepsilon
    },
    \qquad
    j=i-m,\ldots,i-1.
\]
After replaying the stored pairs, we set $\widetilde A_t^{(i)}$ to the resulting
matrix. This corresponds to the \texttt{identity\_rescaled} variant in our
implementation. It is analogous in spirit to limited-memory quasi-Newton methods
such as L-BFGS, where a scalar identity baseline is chosen from recent secant
information and then combined with a finite history of secant updates. In our
experiments, this variant did not lead to significant improvement over simple truncation (Figure \ref{fig:limited}). We note that the L-BFGS-style approach appears slightly more stable than the simple truncation, with a lower third quartile. 

\subsection{Multisecant Broyden update}
\label{app:broyden_multisecant}

The rank-one Broyden update incorporates one secant condition at a time. The
multisecant variant instead fits several recent secant pairs simultaneously. For
each transition $f_t$, let
\[
    P_t^{(i)}
    =
    \left[
        p_t^{(i-m_i)},\ldots,p_t^{(i-1)}
    \right]
    \in \mathbb{R}^{D\times m_i},
    \qquad
    Q_t^{(i)}
    =
    \left[
        q_t^{(i-m_i)},\ldots,q_t^{(i-1)}
    \right]
    \in \mathbb{R}^{D\times m_i},
\]
where $m_i=\min(m,i)$ is the number of stored secants. The multisecant condition
is
\begin{equation}
    \widetilde A_t^{(i)} P_t^{(i)}
    =
    Q_t^{(i)}.
    \label{eq:app_multisecant_condition}
\end{equation}

Unlike the rank-one update, which modifies the previous approximation
$\widetilde A_t^{(i-1)}$, our accelerated variant builds a fresh rank-$m$
correction on top of a diagonal baseline. Let $D_t^{(i)}$ denote this baseline,
for example $D_t^{(i)}=\gamma I_D$ or a freshly computed diagonal approximation.
We write
\[
    \widetilde A_t^{(i)}
    =
    D_t^{(i)}+\Delta_t^{(i)}.
\]
The correction is chosen to have minimum Frobenius norm while satisfying the
stored secant equations:
\[
    \Delta_t^{(i)}
    =
    \arg\min_{\Delta}\|\Delta\|_F
    \quad
    \text{s.t.}
    \quad
    \left(D_t^{(i)}+\Delta\right)P_t^{(i)}=Q_t^{(i)}.
\]
Equivalently,
\[
    \Delta_t^{(i)}P_t^{(i)}
    =
    Q_t^{(i)}-D_t^{(i)}P_t^{(i)}.
\]
Define the residual matrix
\[
    R_t^{(i)}
    :=
    Q_t^{(i)}-D_t^{(i)}P_t^{(i)}
    \in \mathbb{R}^{D\times m_i}.
\]
The minimum-Frobenius-norm solution is
\[
    \Delta_t^{(i)}
    =
    R_t^{(i)}{P_t^{(i)}}^\dagger,
\]
where ${P_t^{(i)}}^\dagger$ is the Moore--Penrose pseudoinverse. If
$P_t^{(i)}$ has full column rank, then
\[
    {P_t^{(i)}}^\dagger
    =
    \left({P_t^{(i)}}^\top P_t^{(i)}\right)^{-1}
    {P_t^{(i)}}^\top.
\]
Thus, in the full-rank case,
\begin{equation}
    \boxed{
    \widetilde A_t^{(i)}
    =
    D_t^{(i)}
    +
    \left(Q_t^{(i)}-D_t^{(i)}P_t^{(i)}\right)
    \left({P_t^{(i)}}^\top P_t^{(i)}\right)^{-1}
    {P_t^{(i)}}^\top
    }
    \label{eq:app_multisecant_update}
\end{equation}
In practice, we use a small ridge term for numerical stability,
\begin{equation}
    \widetilde A_t^{(i)}
    =
    D_t^{(i)}
    +
    \left(Q_t^{(i)}-D_t^{(i)}P_t^{(i)}\right)
    \left({P_t^{(i)}}^\top P_t^{(i)}+\varepsilon I\right)^{-1}
    {P_t^{(i)}}^\top,
    \label{eq:app_multisecant_damped}
\end{equation}
with $\varepsilon>0$ small.

The correction in \eqref{eq:app_multisecant_update} has rank at most $m_i$.
Therefore the accelerated multisecant update produces a DPLR approximation
directly:
\[
    \widetilde A_t^{(i)}
    =
    D_t^{(i)}
    +
    U_t^{(i)}{V_t^{(i)}}^\top,
    \qquad
    \operatorname{rank}\!\left(U_t^{(i)}{V_t^{(i)}}^\top\right)\leq m_i.
\]
With ridge stabilization, one can take
\[
    U_t^{(i)}
    =
    Q_t^{(i)}-D_t^{(i)}P_t^{(i)},
    \qquad
    {V_t^{(i)}}^\top
    =
    \left({P_t^{(i)}}^\top P_t^{(i)}+\eta I\right)^{-1}
    {P_t^{(i)}}^\top .
\]
This is the update used in Accelerated Parallel Broyden: rather than
accumulating one rank-one correction per iteration, it fits several recent
secant directions at once, yielding a rank-at-most-$m$ correction in a single
update.

\subsection{Damping of the low-rank corrections}
\label{app:broyden_damping}

As discussed by \citet{gonzalez2024towards}, large eigenvalues in the linearized
transition dynamics can lead to unstable parallel Newton updates. Their
scale-ELK method addresses this issue by shrinking the full dynamics Jacobian by
a factor $(1-k)$, where $k\in[0,1]$ is a damping parameter. We use the same idea,
but apply it only to the low-rank Broyden correction rather than to the entire
Jacobian approximation.

For the rank-one Broyden update, let
\[
    \Delta_t^{(i)}
    =
    \frac{
        \left(q_t^{(i)}-\widetilde A_t^{(i)}p_t^{(i)}\right)
        {p_t^{(i)}}^\top
    }{
        {p_t^{(i)}}^\top p_t^{(i)}
    }
\]
denote the secant correction. Instead of applying the full correction, we use
\begin{equation}
    \widetilde A_t^{(i+1)}
    =
    \widetilde A_t^{(i)}
    +
    (1-k)\Delta_t^{(i)},
    \qquad
    k \in [0,1].
    \label{eq:app_damped_broyden}
\end{equation}
Thus $k=0$ recovers the standard Broyden update, while larger $k$ values damp
the new low-rank contribution. At $k=1$, the secant correction is completely
suppressed and the block approximation is left unchanged.

For the multisecant update, recall that Accelerated Parallel Broyden builds a
fresh correction on top of a diagonal baseline $D_t^{(i)}$. With damping, we use
\begin{equation}
    \widetilde A_t^{(i)}
    =
    D_t^{(i)}
    +
    (1-k)
    \left(Q_t^{(i)}-D_t^{(i)}P_t^{(i)}\right)
    \left({P_t^{(i)}}^\top P_t^{(i)}+\eta I\right)^{-1}
    {P_t^{(i)}}^\top .
    \label{eq:app_damped_multisecant}
\end{equation}
Here $P_t^{(i)}$ and $Q_t^{(i)}$ are the matrices of recent trajectory secants
and dynamics finite differences, respectively, and $\eta>0$ is the small ridge
parameter used for numerical stability.

\begin{figure}
    \centering
    \includegraphics[width=\linewidth]{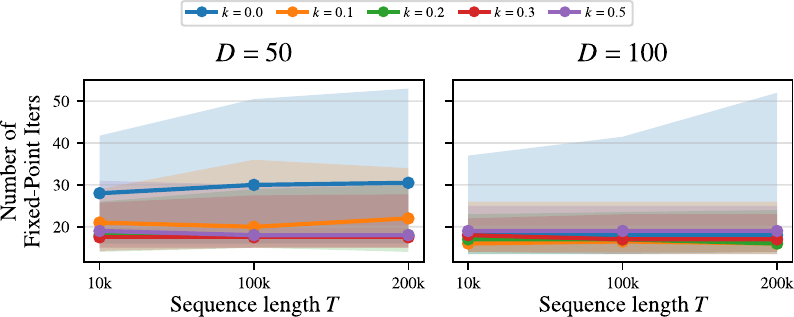}
    \caption{\textbf{Impact of damping on Parallel Broyden stability.}
Low-rank RNN experiment over $32$ random seeds. Points denote medians and error
bars denote interquartile ranges. Moderate damping reduces instability and
improves robustness across seeds, while excessive damping weakens the low-rank
secant correction and can degrade convergence. For example, when $D=100$,
$k=0.5$ gives worse median performance than smaller damping values across all
tested trajectory lengths.}
    \label{fig:app_damping}
\end{figure}

The damping parameter controls a stability--accuracy tradeoff. Larger values of
$k$ shrink the low-rank correction and can reduce instabilities caused by large
eigenvalues in the approximate transition blocks. However, excessive damping
also weakens the secant information, making the Jacobian approximation closer to
its diagonal baseline and typically increases the number of Newton iterations (Figure \ref{fig:app_damping}).
In practice, we therefore choose $k$ heuristically as small as possible while
remaining stable across the tested instances. We tune $k$ per method and setting. The values are given in Appendices~\ref{app:low_rank_rnn} and~\ref{app:gibbs}. We did not need damping for the diffusion experiment.

\section{RNN experiment details}
\label{app:rnn}

All RNN experiments use $32$ chains, a maximum of $1000$ iterations, and the stopping criterion
\[
    \frac{\|\mathbf{r}(\mathbf{x}_{1:T})\|_2}{\sqrt{TD}} < 10^{-4},
\]
that is, an absolute root-mean-square one-step residual below $10^{-4}$ per state coordinate, which is small relative to the scale of the states. Computations are in \texttt{float32} with highest-precision matrix multiplications. We report medians and interquartile ranges over chains. Iteration counts include all chains, with non-converged chains counted at the $1000$-iteration cap, and runtimes are computed over converged chains. The noise sequence $\{\epsilon_t\}_{t=1}^T$ is fixed before running the solver, so each model defines a deterministic input-driven state-space system. Newton uses the full Jacobian computed by reverse-mode automatic differentiation, and diagonal quasi-Newton extracts its diagonal from it. For the models considered here, the Jacobian and its diagonal are available in closed form or cheaply from the low-rank structure, and would be computed directly in practice. We nevertheless treat the transition $f$ as a black box and obtain them by automatic differentiation, as one would for a general nonlinear state-space model. For both Broyden variants, the low-rank products formed during the parallel scan are compressed back to rank $m$ with a rank-revealing variant of the randomized compression described in the main text. In this variant, sketch columns that are numerically dependent, with a relative remainder below $10^{-4}$, are dropped rather than normalized. This prevents rank-deficient corrections, which are frequent in \texttt{float32}, from introducing spurious directions. Accelerated Parallel Broyden uses a Gram regularization $\varepsilon=10^{-6}$. All runtimes are measured on a single NVIDIA H100 GPU.

\subsection{Low-rank RNNs}
\label{app:low_rank_rnn}

The state $x_t \in \mathbb{R}^D$ evolves according to
\begin{equation}
    x_t
    =
    a x_{t-1}
    +
    \frac{\Delta t}{\tau} M \psi\!\left(N^\top \phi(x_{t-1})\right)
    +
    \epsilon_t,
    \qquad
    M,N \in \mathbb{R}^{D \times R},
    \label{eq:app_low_rank_rnn}
\end{equation}
where $a = 1-\Delta t/\tau$, $\phi=\tanh$ is applied elementwise, and $\epsilon_t \sim \mathcal{N}(0,\sigma^2 I_D)$ with $\sigma=0.1$. The standard low-rank RNN uses $\psi(z)=z$, so that the recurrent weight matrix is $W=MN^\top$. The low-rank MLP-RNN uses $\psi(z)=z+\beta\,\mathrm{MLP}(z)$ with $\beta=0.5$, where the MLP has $8$ hidden layers of width $D$ and tanh activations. Given a gain $g$, we draw
\begin{equation}
    M_{ij} \sim \mathcal{N}\!\left(0,\frac{g^2}{D}\right),
    \qquad
    N_{ij} \sim \mathcal{N}\!\left(0,\frac{1}{R}\right).
\end{equation}
This normalization keeps the coordinate-wise recurrent input at scale $O(g)$ when the components of $\phi(x)$ are $O(1)$, while the nonzero singular values of $MN^\top$ are typically of order $g\sqrt{D/R}$. Each chain draws its own weights, noise, initial state and initial guess, with $x_0, x^{(0)}_{1:T}\sim 0.1\,\mathcal{N}(0,I)$. Seeds whose sequential trajectory saturates are discarded. The criteria chosen for saturation are that, after a burn-in of $T/4$ steps, some unit has mean $|\tanh|>0.8$ or mean $\tanh'<0.1$.

\begin{figure}[t]
    \centering
    \includegraphics[width=0.4\linewidth]{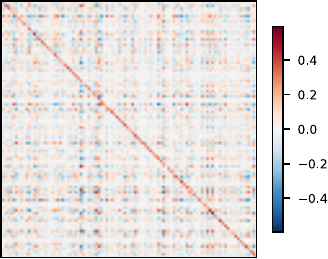}
    \caption{\textbf{Example of the dynamics Jacobian matrix of the standard low-rank RNN.} The matrix is exactly diagonal plus rank $R=3$.}
    \label{fig:app_low_rank_jac}
\end{figure}

The transition Jacobian of \eqref{eq:app_low_rank_rnn} is
\begin{equation}
    A_t
    =
    \frac{\partial x_t}{\partial x_{t-1}}
    =
    a I_D
    +
    \frac{\Delta t}{\tau}
    M\, \psi'\!\left(N^\top\phi(x_{t-1})\right) N^\top \operatorname{diag}\!\left(\phi'(x_{t-1})\right),
    \label{eq:app_low_rank_rnn_jac}
\end{equation}
where $\psi'$ is the $R\times R$ Jacobian of $\psi$. For both models, the off-diagonal term has rank at most $R$, so the true Jacobian is exactly diagonal plus rank $R$ (Figure~\ref{fig:app_low_rank_jac}). The MLP makes each evaluation of $f$ and each Jacobian computation more expensive without changing this structure.

We use $R=3$, $g=2$, $\tau=10$, and $D\in\{100,200\}$, with $D=200$ in Figure~\ref{fig:fig1}. The step size $\Delta t$ controls the balance between the two terms of \eqref{eq:app_low_rank_rnn_jac}. Before modulation by $\psi'$ and $\phi'$, the low-rank term has typical spectral scale $(\Delta t/\tau)\, g\sqrt{D/R}$, which is about $0.8$ for $D=200$ and $\Delta t=0.5$, against a diagonal $a=0.95$. For $\Delta t=3$ it is about $4.9$, against $a=0.7$. The off-diagonal coupling is therefore weak for $\Delta t=0.5$ and dominant for $\Delta t=3$. Table~\ref{tab:app_lowrank_ratio} reports iteration counts for both models and both step sizes. In all settings, Accelerated Parallel Broyden needs fewer iterations than Parallel Broyden, and the gap is larger for $\Delta t=3$, independently of the model.

For Parallel Broyden and Accelerated Parallel Broyden, we use rank $m=R=3$ and initialize the diagonal at $\gamma=a$, so that $\widetilde{A}_t^{(0)}=aI_D$ matches the diagonal of \eqref{eq:app_low_rank_rnn_jac}. Accelerated Parallel Broyden uses a backtracking line search with sufficient-decrease constant $c=10^{-3}$, and Parallel Broyden uses none. For each method, the damping minimizes the median iteration count over chains, with non-converged chains counted at the cap. It is tuned on $T\in\{10^4,10^5\}$, or on $T=10^4$ only for diagonal quasi-Newton, and reused for $T=10^3$. The resulting values are listed in Table~\ref{tab:app_lowrank_damping}.

\begin{table}[h]
    \centering
    \caption{\textbf{Damping values for the low-rank RNNs.} The same values are used for both models.}
    \label{tab:app_lowrank_damping}
    \begin{tabular}{lcc}
        \toprule
        Method & $\Delta t=0.5$ & $\Delta t=3$ \\
        \midrule
        Newton & $0$ & $0$ \\
        Diag.\ quasi-Newton & $0.1$ & $0.8$ \\
        Parallel Broyden & $0.2$ & $0.4$ \\
        Accel.\ Parallel Broyden & $0.2$ & $0.1$ \\
        \bottomrule
    \end{tabular}
\end{table}

\begin{table}[h]
    \centering
    \caption{\textbf{Iterations of the Broyden variants on the low-rank RNNs} ($D=200$). Median over all $32$ chains, with non-converged chains counted at the $1000$-iteration cap, and number of converged chains.}
    \label{tab:app_lowrank_ratio}
    \begin{tabular}{llrrrrrr}
        \toprule
        & & & \multicolumn{2}{c}{Iterations} & & \multicolumn{2}{c}{Converged} \\
        \cmidrule(lr){4-5} \cmidrule(lr){7-8}
        Model & $\Delta t$ & $T$ & PB & Accel.\ PB & PB / Accel. & PB & Accel.\ PB \\
        \midrule
        Standard & $0.5$ & $10^3$ & $12.5$ & $10$   & $1.25$ & $32$ & $32$ \\
                 &       & $10^4$ & $14$   & $10$   & $1.40$ & $30$ & $32$ \\
                 &       & $10^5$ & $15$   & $12$   & $1.25$ & $30$ & $32$ \\
                 & $3$   & $10^3$ & $30.5$ & $10$   & $3.05$ & $32$ & $32$ \\
                 &       & $10^4$ & $29.5$ & $11$   & $2.68$ & $27$ & $31$ \\
                 &       & $10^5$ & $30$   & $12$   & $2.50$ & $26$ & $30$ \\
        \midrule
        MLP      & $0.5$ & $10^3$ & $12$   & $9.5$  & $1.26$ & $32$ & $32$ \\
                 &       & $10^4$ & $15$   & $10$   & $1.50$ & $30$ & $32$ \\
                 &       & $10^5$ & $15$   & $12.5$ & $1.20$ & $30$ & $32$ \\
                 & $3$   & $10^3$ & $28.5$ & $10.5$ & $2.71$ & $32$ & $32$ \\
                 &       & $10^4$ & $28$   & $11.5$ & $2.43$ & $27$ & $31$ \\
                 &       & $10^5$ & $26.5$ & $12.5$ & $2.12$ & $27$ & $31$ \\
        \bottomrule
    \end{tabular}
\end{table}

\subsection{Full-rank DMF RNN}
\label{app:dmf_rnn}

\begin{figure}[t]
    \centering
    \includegraphics[width=0.5\linewidth]{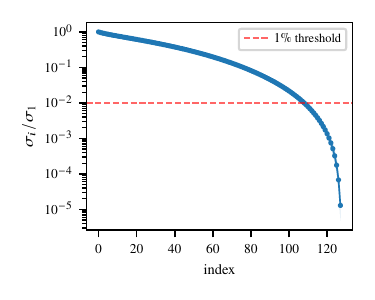}
    \caption{\textbf{Singular value spectrum of the DMF RNN Jacobian.} Normalized singular values $\sigma_i/\sigma_1$ of $A_t$ along trajectories (median and interquartile range over $32$ independently sampled networks). The dashed line marks $1\%$ of $\sigma_1$.}
    \label{fig:app_dmf_spectrum}
\end{figure}

The discretized mean-field RNN~\citep{engelken2023lyapunov} evolves according to
\begin{equation}
    x_t = W \tanh(x_{t-1}) + \epsilon_t,
    \qquad
    W_{ij} \sim \mathcal{N}\!\left(0,\frac{g^2}{D}\right),
\end{equation}
with $g=1.7$, $D=128$, self-couplings included, $\epsilon_t \sim \mathcal{N}(0, I_D)$, and $T=10^4$. A single weight matrix $W$ is shared by all $32$ chains, which differ in their noise sequence and initial guess, with $x_0=0$ and $x^{(0)}_{1:T}\sim\mathcal{N}(0,I)$. The Jacobian $A_t = W\operatorname{diag}(\operatorname{sech}^2(x_{t-1}))$ is dense and full rank, with no dominant diagonal. Along trajectories, its normalized singular values decay slowly, and about $108$ of the $128$ exceed $1\%$ of the largest (Figure~\ref{fig:app_dmf_spectrum}). No small rank $m$ can therefore represent $A_t$ exactly, which makes this model a test of how the Broyden variants degrade as $m$ decreases.

Since the model has no leak term, the Broyden variants initialize the diagonal at $\gamma=0.1$. Both Broyden variants use a backtracking line search with $c=10^{-3}$. We sweep $m\in\{1,5,10,\dots,80\}$ for Parallel Broyden and $m\in\{1,5,10,15,20\}$ for Accelerated Parallel Broyden. For larger $m$, the Gram matrix of Accelerated Parallel Broyden becomes nearly singular, which degrades convergence. Newton uses damping $0.2$, which prevents numerical overflow in \texttt{float32}, and all other methods use no damping.

\section{Hierarchical Gibbs sampler details}
\label{app:gibbs}

We use a Gibbs sampler for a hierarchical normal model based on the eight schools problem, with a variable number of schools $S$ to study scaling with state dimension. For each school $s=1,\ldots,S$ and student $n=1,\ldots,N$, the model is
\begin{align*}
    y_{s,n} \mid \theta_s,\sigma_s^2
    &\sim
    \mathcal{N}(\theta_s,\sigma_s^2),
    \\
    \theta_s \mid \mu,\tau^2
    &\sim
    \mathcal{N}(\mu,\tau^2).
\end{align*}
We place conjugate priors on the global mean, global variance, and school-level variances:
\begin{align*}
    \mu \mid \tau^2
    &\sim
    \mathcal{N}\!\left(\mu_0,\tau^2/\kappa_0\right),
    \\
    \tau^2
    &\sim
    \mathrm{InvGamma}\!\left(\frac{\nu_0}{2}, \frac{\nu_0\tau_0^2}{2}\right),
    \\
    \sigma_s^2
    &\sim
    \mathrm{InvGamma}\!\left(\frac{\alpha_0}{2}, \frac{\alpha_0\sigma_0^2}{2}\right).
\end{align*}
The MCMC state at sweep $t$ is
\[
    x_t =
    \bigl(\theta_{1:S,t}, \sigma^2_{1:S,t}, \mu_t, \tau_t^2\bigr)
    \in \mathbb{R}^{2S+2}.
\]
In the experiments, we set
\[
    N=20,\quad
    \mu_0=0,\quad
    \kappa_0=0.1,\quad
    \nu_0=0.1,\quad
    \tau_0^2=100,\quad
    \alpha_0=0.1,\quad
    \sigma_0^2=10.
\]
We generate synthetic school-level summaries by drawing
\[
    \bar{y}_s \sim \mathcal{N}(0,10^2),
    \qquad
    \bar{\sigma}_s = |\eta_s|+5,
    \qquad
    \eta_s\sim\mathcal{N}(0,5^2).
\]
Given these summaries, we generate observations by first sampling
\[
    y^{\mathrm{raw}}_{s,n}
    \sim
    \mathcal{N}\!\left(\bar{y}_s,N\bar{\sigma}_s^2\right),
\]
then standardizing each school's column to $z_{s,n}$ and setting $y_{s,n}=\bar y_s+\sqrt{N}\,\bar\sigma_s z_{s,n}$. Each school's sample mean is then exactly $\bar y_s$ and its sample standard deviation $\sqrt{N}\bar\sigma_s$, so that its standard error is $\bar\sigma_s$.

\paragraph{Conditional updates.}
Each Gibbs sweep updates $\tau^2$, then $\mu$, then $\theta_{1:S}$, then $\sigma^2_{1:S}$. Let
\[
    \nu_N=\nu_0+S+1,
    \qquad
    \alpha_N=\alpha_0+N.
\]
Given independent random variates fixed by the driver at sweep $t$, the updates are
\begin{align*}
    \tau_N^2
    &=
    \frac{1}{\nu_N}
    \left[
        \nu_0\tau_0^2
        + \kappa_0(\mu-\mu_0)^2
        + \sum_{s=1}^S(\theta_s-\mu)^2
    \right],
    \\
    \tau^2
    &=
    \frac{\nu_N \tau_N^2}{2}\,\xi_\tau,
    \qquad
    \xi_\tau \sim \mathrm{InvGamma}\!\left(\frac{\nu_N}{2},1\right),
    \\
    v_\mu
    &=
    \frac{\tau^2}{\kappa_0+S},
    \qquad
    \hat{\mu}
    =
    \frac{\kappa_0\mu_0+\sum_{s=1}^S\theta_s}{\kappa_0+S},
    \\
    \mu
    &=
    \hat{\mu}
    +
    \sqrt{v_\mu}\,\xi_\mu,
    \qquad
    \xi_\mu\sim\mathcal{N}(0,1),
    \\
    v_{\theta_s}
    &=
    \left(\frac{N}{\sigma_s^2}+\frac{1}{\tau^2}\right)^{-1},
    \qquad
    \hat{\theta}_s
    =
    v_{\theta_s}
    \left(
        \frac{\sum_{n=1}^{N}y_{s,n}}{\sigma_s^2}
        +
        \frac{\mu}{\tau^2}
    \right),
    \\
    \theta_s
    &=
    \hat{\theta}_s+\sqrt{v_{\theta_s}}\,\xi_{\theta_s},
    \qquad
    \xi_{\theta_s}\sim\mathcal{N}(0,1),
    \\
    \sigma_{N,s}^2
    &=
    \frac{1}{\alpha_N}
    \left[
        \alpha_0\sigma_0^2
        +
        \sum_{n=1}^{N}(y_{s,n}-\theta_s)^2
    \right],
    \\
    \sigma_s^2
    &=
    \frac{\alpha_N\sigma_{N,s}^2}{2}\,\xi_{\sigma_s},
    \qquad
    \xi_{\sigma_s}
    \sim
    \mathrm{InvGamma}\!\left(\frac{\alpha_N}{2},1\right).
\end{align*}
Fixing the random variates
\[
    \xi_t =
    \left(
    \xi_{\theta,1:S,t},
    \xi_{\sigma,1:S,t},
    \xi_{\mu,t},
    \xi_{\tau,t}
    \right)
\]
for every sweep turns the Gibbs sampler into a deterministic input-driven state-space model,
\[
    x_{t+1}=f_t(x_t;\xi_t).
\]

\begin{figure}
    \centering
    \includegraphics[width=\linewidth]{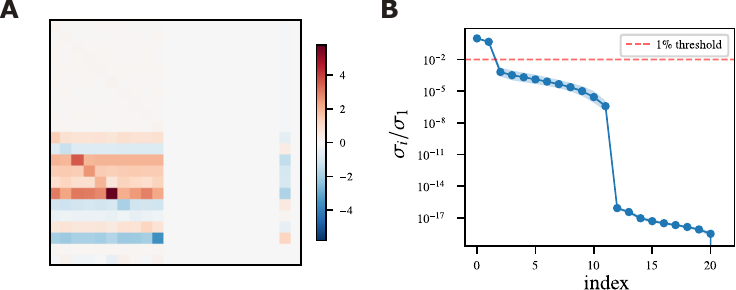}
    \caption{\textbf{Structure of the Gibbs transition Jacobian for the hierarchical normal model.} \textbf{A:} Example transition Jacobian averaged over time, showing small diagonal entries and structured off-diagonal coupling induced by the hierarchical updates. \textbf{B:} Singular values of the transition Jacobian, normalized by the largest singular value and averaged over time, with error bars denoting standard deviations. The spectrum is dominated by the first two singular values, supporting the rank-$2$ purely low-rank approximation used by the Broyden variants in this experiment.}
    \label{fig:mcmc_jac}
\end{figure}

\paragraph{Why the Jacobian is approximately low rank.}
The hierarchical structure induces strong cross-school coupling through the two global variables $\mu$ and $\tau^2$. The school-level variables affect the global updates through sums such as $\sum_s \theta_s$ and $\sum_s(\theta_s-\mu)^2$, while the updated global variables are then broadcast back to all school-level updates through the conditional means and variances of $\theta_s$. Thus, the dominant off-diagonal interactions are mediated by a small number of global degrees of freedom. This motivates a rank-$2$ approximation corresponding to the two global variables $(\mu,\tau^2)$ (Figure~\ref{fig:mcmc_jac}).

At the same time, the diagonal entries of the transition Jacobian are small, because Gibbs updates resample each coordinate from its conditional distribution. Each updated coordinate depends primarily on the other variables that determine its conditional distribution, rather than on its own previous value. For this reason, both Broyden variants initialize the diagonal term to zero ($\gamma=0$) and use rank $m=2$. Since the approximation is then purely low rank, products of rank-$m$ blocks remain rank at most $m$, and no rank compression is required.

\paragraph{Experimental setup.}
We use $S\in\{10,50,100,200\}$, so that $D=2S+2$ ranges from $22$ to $402$, and $T\in\{10^3,10^4,5\cdot10^4,10^5\}$. Each setting uses $32$ chains. Each chain has its own synthetic dataset and its own random variates $\xi_{1:T}$. Its initial state is drawn from the prior and followed by one Gibbs sweep, and the initial guess repeats this state at every time step. All methods use a maximum of $1000$ iterations, computations in \texttt{float32} with highest-precision matrix multiplications, and the stopping criterion
\[
    \frac{\|\mathbf{r}(\mathbf{x}_{1:T})\|_2}{\sqrt{TD}} < 10^{-3},
\]
a tolerance that is small relative to the scale of the Gibbs state variables. We report medians and interquartile ranges over all chains, with non-converged chains counted at the $1000$-iteration cap and at their actual runtime. Newton uses the full Jacobian computed by reverse-mode automatic differentiation, and diagonal quasi-Newton extracts its diagonal from it. Newton and diagonal quasi-Newton use no damping, Parallel Broyden uses damping $0.3$, and Accelerated Parallel Broyden uses damping $0.1$. Accelerated Parallel Broyden also uses a backtracking line search with sufficient-decrease constant $c=10^{-3}$ and a Gram regularization $\varepsilon=10^{-6}$, while Parallel Broyden uses no line search. After each iteration, both Broyden variants clip the variances $\sigma_s^2$ and $\tau^2$ to at least $10^{-6}$ to keep the iterates in the support of the model. For $S=200$ and $T\geq 5\cdot10^4$, Newton runs out of memory. All runtimes are measured on a single NVIDIA H100 GPU.

\paragraph{Stochastic diagonal estimation.}
Replacing the exact diagonal of $A_t$ with a Hutchinson estimate, as in the diffusion experiment, breaks diagonal quasi-Newton on this benchmark (Table~\ref{tab:nschools_hutchinson}). With $1$ or $8$ Rademacher probes per step, it fails on every chain. With $64$ probes, some chains converge at $T=10^3$, but only after $963$ to $999$ iterations, close to the worst-case bound of $T$. By comparison, at $T=10^3$ the median iteration count is $68$ to $136$ with the exact diagonal and $21$ to $32$ for Accelerated Parallel Broyden. This failure is expected from the structure of $A_t$: since its diagonal entries are small, the off-diagonal entries carry most of its Frobenius norm, and the noise of the estimate is large relative to the diagonal it estimates.

\begin{table}[h]
    \centering
    \caption{\textbf{Hutchinson-based diagonal quasi-Newton on the hierarchical Gibbs sampler.} Number of chains, out of $32$, that reach the tolerance within $1000$ iterations.}
    \label{tab:nschools_hutchinson}
    \begin{tabular}{ccrrrr}
        \toprule
        Probes & $T$ & $S=10$ & $S=50$ & $S=100$ & $S=200$ \\
        \midrule
        $1$  & $10^3$ & $0$ & $0$  & $0$  & $0$  \\
        $1$  & $10^4$ & $0$ & $0$  & $0$  & $0$  \\
        $8$  & $10^3$ & $0$ & $0$  & $0$  & $0$  \\
        $8$  & $10^4$ & $0$ & $0$  & $0$  & $0$  \\
        $64$ & $10^3$ & $1$ & $19$ & $22$ & $16$ \\
        $64$ & $10^4$ & $0$ & $0$  & $0$  & $0$  \\
        \bottomrule
    \end{tabular}
\end{table}

\section{Diffusion experiment}
\label{app:ddim}

\subsection{Structure of the local dynamics Jacobian}
\label{app:ddim_jac}

\begin{figure}
    \centering
    \includegraphics[width=0.5\linewidth]{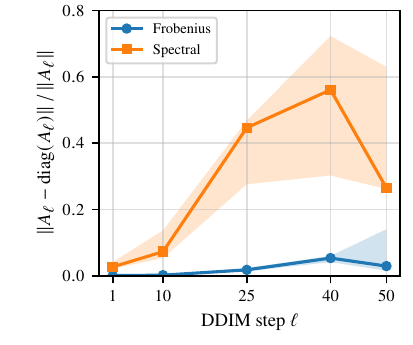}
    \caption{\textbf{Off-diagonal mass of the DDIM local dynamics Jacobian.} Relative norm of $A_\ell-\mathrm{diag}(A_\ell)$ along the sampling trajectory, in Frobenius and spectral norm (median and range over $3$ images).}
    \label{fig:dit_jacobian}
\end{figure}

Recall that the local dynamics Jacobian of the DDIM update is
\[
    A_\ell(x_{\ell-1})
    =
    c_x(t)I
    +
    c_\epsilon(t)\frac{\partial \epsilon_\theta}{\partial x}(x_{\ell-1},t),
    \qquad t = T-\ell+1,
\]
a $4096\times4096$ matrix. To quantify how well a diagonal approximation captures it, we form $A_\ell$ exactly by automatic differentiation through the DiT denoiser along the DDIM sampling trajectory. We do this for $\ell\in\{1,10,25,40,50\}$ and three class-conditional images, using the same pretrained model, guidance scale and $T=50$ as in the main experiments. Figure~\ref{fig:dit_jacobian} reports the relative off-diagonal mass $\|A_\ell-\mathrm{diag}(A_\ell)\|/\|A_\ell\|$ in Frobenius and spectral norm.

In Frobenius norm, $A_\ell$ is close to diagonal throughout sampling. Its off-diagonal part accounts for under $1\%$ of $\|A_\ell\|_F$ early in sampling and at most about $5\%$ in the median at later steps, with one image reaching $14\%$ at the final step. The spectral ratio is larger: it rises after mid-sampling and peaks at about $0.5$ around $\ell=40$. This indicates that the off-diagonal coupling is small in total but concentrated in a few directions of the state, where the diagonal approximation is inaccurate. Such an error mainly slows the corresponding modes of the iteration, which is consistent with diagonal quasi-Newton requiring only slightly more iterations than Parallel Broyden in Figure~\ref{fig:fig3}. The small Frobenius ratio also explains why a single-probe Hutchinson estimate is accurate on this benchmark. For a Rademacher probe $z$, the error of the estimate $z\odot(A_\ell z)$ comes only from the off-diagonal entries of $A_\ell$, and its root-mean-square relative error is $\|A_\ell-\mathrm{diag}(A_\ell)\|_F/\|\mathrm{diag}(A_\ell)\|_F$, which nearly coincides with the Frobenius curve in Figure~\ref{fig:dit_jacobian} since the diagonal dominates.

\subsection{Hyperparameter setting for ParaTAA}
\label{app:parataa_hparams}

ParaTAA~\citep{tang2024accelerating} parallelizes diffusion sampling by rewriting the denoising trajectory as a triangular nonlinear system and applying triangular Anderson acceleration. The method has three relevant hyperparameters: the window size $w$, the order $k$ of the equivalent triangular system, and the history size $m$. In our DDIM-50 experiment, we follow ParaTAA and set the window size equal to the number of sampling steps, $w=T=50$, so the full trajectory is solved in a single window. The order $k$ controls how many future timesteps are used in each triangular update, while $m$ controls the amount of Anderson history used to build the inverse multisecant approximation.

\citet{tang2024accelerating} treat $k$ and $m$ as empirical hyperparameters and report two trends relevant to our sweep: $k$ should be tuned because it controls a stability--information-propagation tradeoff, and the history size is typically kept small because additional history does not necessarily improve convergence and can hurt performance. We therefore perform a small grid search over $k$ and $m$ in our DDIM-50 setting.

\begin{figure}
    \centering
    \includegraphics[width=0.55\linewidth]{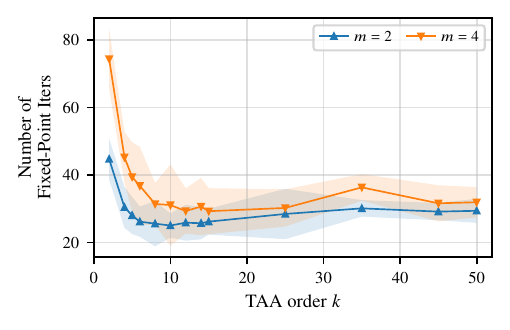}
    \caption{\textbf{ParaTAA hyperparameter search.} We vary the nonlinear system order $k$ and history size $m$ for DDIM-50 with window size $w=T=50$. Points denote means over $32$ images, and shaded regions denote $\pm$ one standard deviation. Consistent with ParaTAA's hyperparameter analysis, small history sizes are sufficient: $m=2$ requires fewer iterations than $m=4$ at every order.}
    \label{fig:taa_search}
\end{figure}

Figure~\ref{fig:taa_search} shows that both hyperparameters matter in our setting. Small orders propagate information slowly: at $k=2$, the mean iteration count is $44.7$ for $m=2$ and $74.3$ for $m=4$. For $m=2$, performance improves rapidly with $k$ and is nearly flat for $6\le k\le 15$ (mean $25.0$--$26.2$ iterations and mean runtime $3.2$--$3.4$\,s), before degrading slightly for larger orders (mean $29.2$--$30.1$ iterations for $k\ge 35$), consistent with the stability vs. information-propagation tradeoff described by \citet{tang2024accelerating}. Increasing the history size to $m=4$ is worse than $m=2$ at every order. Based on this sweep, we use $m=2$ and $k=10$ for the ParaTAA baseline, which attains the lowest mean iteration count ($25.0$). With $k=8$, it is nearly as good on average ($25.6$) but has a much worse worst case ($62$ vs.\ $41$ iterations).

\subsection{Hyperparameter setting for Picard iteration}
\label{app:paradigms_hparams}

ParaDiGMS~\citep{shih2023parallel} parallelizes diffusion sampling with Picard iteration. In our notation, this corresponds to a fixed-point parallel trajectory update, or equivalently to using the identity local approximation $\widetilde A_\ell=I_D$. Unlike Parallel Broyden, ParaDiGMS does not build secant-based Jacobian approximations. It repeatedly refines the full denoising trajectory using parallel denoiser evaluations.

The main hyperparameter is the sliding-window size $w$, which controls how many timesteps are updated in parallel. The original method uses this window to manage memory when the full trajectory is too large to evaluate at once. In our DDIM-50 experiments, we set $w=T=50$, so the full trajectory is updated in a single window. For SDE samplers, ParaDiGMS samples the noise up front and treats it as part of a deterministic input-driven trajectory. Since we evaluate DDIM here, no additional noise is injected after the initial latent is fixed.

\subsection{Setting for diagonal quasi-Newton}
\label{app:quasi_hparams}

Diagonal quasi-Newton estimates the diagonal of each $A_\ell$ with the Hutchinson estimator using a single Rademacher probe $z$, $\mathrm{diag}(A_\ell)\approx z\odot(A_\ell z)$, which requires one Jacobian--vector product through the denoiser per time step and iteration. Forming the exact diagonal would require the full $4096\times4096$ Jacobian. As shown in Appendix~\ref{app:ddim_jac}, a single probe is accurate on this benchmark because $A_\ell$ is close to diagonal in Frobenius norm.

\subsection{Hyperparameter setting for Parallel Broyden}
\label{app:broyden_hparams}

Parallel Broyden has two main hyperparameters in the DDIM experiment: the maximum DPLR rank $m$, which controls the number of stored Broyden corrections, and the diagonal scaling $\gamma$ in the initialization $\gamma I_D$. We tune these hyperparameters with a small grid search on the DDIM-50 benchmark, using the same $32$ images as in the main evaluation. We use the same selected hyperparameters for Accelerated Parallel Broyden.

\begin{figure}
    \centering
    \includegraphics[width=0.55\linewidth]{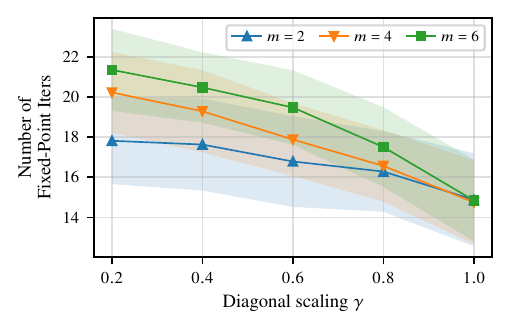}
    \caption{\textbf{Parallel Broyden hyperparameter search.} We vary the diagonal scaling $\gamma$ in the initialization $\gamma I_D$ and the maximum DPLR rank $m$ on the DDIM-50 benchmark. Points denote means over $32$ images, and shaded regions denote $\pm$ one standard deviation. Larger diagonal scaling improves convergence across all tested ranks, while increasing the rank beyond $m=2$ gives no improvement.}
    \label{fig:broyden_hparams}
\end{figure}

Figure~\ref{fig:broyden_hparams} shows that the best performance is obtained with $\gamma=1.0$. For each rank, increasing $\gamma$ decreases the mean iteration count and runtime, suggesting that the identity-scaled diagonal term already captures an important part of the DDIM local dynamics. Increasing the rank from $m=2$ to $m=4$ or $m=6$ does not meaningfully improve the best mean iteration count. Based on this sweep, we use $\gamma=1.0$ and $m=2$ for both Parallel Broyden and Accelerated Parallel Broyden in the DDIM experiments.

Both Broyden variants use no damping and no line search. The low-rank products formed during the parallel scan are compressed with the randomized compression described in the main text. The rank-revealing variant used in the other benchmarks is not needed here. Accelerated Parallel Broyden uses a Gram regularization $\varepsilon=10^{-10}$.

\subsection{Image quality and stopping criterion}
\label{app:ddim_quality}

All DDIM experiments in this work are run on a single NVIDIA H100 GPU in \texttt{float32}. Since the trajectory is short ($T=50$), default matrix-multiplication precision suffices, unlike in the other benchmarks. The only exception is ParaTAA, which accumulates and solves its Anderson Gram system in \texttt{float64} with Tikhonov regularization $\lambda=10^{-3}$, following the reference implementation~\citep{tang2024accelerating}. Since parallel diffusion samplers solve the sampling trajectory only approximately, we need both a stopping criterion and a way to check that the resulting images match the sequential ones.

\paragraph{Image-quality metric.}
We compare each parallel sample to the corresponding sequential DDIM sample using LPIPS, or Learned Perceptual Image Patch Similarity~\citep{zhang2018unreasonable}. LPIPS measures perceptual distance by comparing deep feature activations between two images, and is often preferred to pixelwise metrics such as mean squared error because it correlates better with human perceptual judgments. Lower LPIPS values indicate greater perceptual similarity, with 0 corresponding to identical images in the learned feature space. All LPIPS values are computed over the same 32 images used for the runtime and iteration-count experiments.

\paragraph{Stopping criterion.}
For all methods we use the same criterion: we stop the trajectory solver when $\|r(x_{1:T})\|_2^2/2 < 0.1$, where $r$ is the single-step DDIM residual in latent space, with a maximum of $100$ iterations. Using a common criterion makes iteration counts directly comparable across methods. All hyperparameter sweeps use the same criterion.

We chose this threshold with a tolerance study: each method was run for a fixed $50$ iterations, and we recorded the residual and the LPIPS distance to the sequential sample along the trajectory. At a tolerance of $0.1$, the image quality of all methods is essentially at convergence: for every image, the solver has already achieved at least $96\%$ of the total LPIPS reduction observed between iteration $3$ and iteration $50$, and for $31$ or $32$ of the $32$ images per method more than $99\%$. Consistent with this, we could not visually distinguish samples obtained at this tolerance from fully converged ones, and tightening the tolerance further yields only marginal gains while adding iterations. A looser tolerance of $1$ stops some trajectories too early: only about $22$ to $28$ of the $32$ images per method have achieved more than $99\%$ of their LPIPS reduction. The stricter tolerance costs a median of $2$ additional iterations for the Broyden, Picard and quasi-Newton methods and $5$ for ParaTAA. Using \texttt{float64} for ParaTAA's Gram system leaves its median iteration count unchanged and mainly shortens one slow case ($52 \to 41$ iterations for the sea turtle image).

This choice follows the principle noted by ParaTAA~\citep{tang2024accelerating}: generated images can appear visually satisfactory before the nonlinear residual reaches a strict numerical tolerance, so it is unnecessary to solve the full trajectory to machine precision.

\paragraph{Resulting image quality.}
Table~\ref{tab:ddim_lpips} reports the LPIPS distances between the parallel samples, obtained with this stopping criterion, and the corresponding sequential DDIM samples. All methods achieve very small LPIPS values (medians of about 0.003), indicating that the images are perceptually very close to the sequential rollout and, in practice, indistinguishable by visual inspection. The two larger values (up to about 0.07) occur for the same images under every method and persist after full convergence. We report medians and interquartile ranges for LPIPS, unlike the mean-based Figures~\ref{fig:fig3}, \ref{fig:taa_search} and~\ref{fig:broyden_hparams}, because these two images inflate both the mean and the standard deviation: for every method, the standard deviation is about twice the mean, so a mean~$\pm$~standard deviation summary would not describe the typical image.

\begin{table}[t]
\centering
\caption{LPIPS distance between each parallel sample and the corresponding sequential DDIM reference over $32$ images. Lower is better.}
\label{tab:ddim_lpips}
\begin{tabular}{lcc}
\toprule
Method & Median LPIPS & IQR \\
\midrule
Diagonal quasi-Newton & $0.0032$ & $[0.0014,\ 0.0065]$ \\
Parallel Broyden & $0.0032$ & $[0.0015,\ 0.0051]$ \\
Picard & $0.0034$ & $[0.0015,\ 0.0055]$\\
Accelerated Parallel Broyden & $0.0031$ & $[0.0014,\ 0.0051]$ \\
ParaTAA & $0.0037$ & $[0.0016,\ 0.0062]$ \\
\bottomrule
\end{tabular}
\end{table}

\section{Runtime decomposition}
\label{app:runtime}

The runtime ratio between a baseline and a Broyden variant factors into an iteration ratio, which reflects the quality of the Jacobian approximation, and a per-iteration cost ratio, which reflects the cost of obtaining it (Table~\ref{tab:decomposition}). The DPLR blocks always need fewer iterations than diagonal quasi-Newton, from $1.1\times$ fewer on DiT to $5.6\times$ fewer on the Gibbs sampler. Most of the speedup, however, comes from avoiding automatic differentiation. On the MLP-RNN and the Gibbs sampler, diagonal quasi-Newton with the exact diagonal spends $91$--$99\%$ of each iteration in automatic differentiation (Table~\ref{tab:wallclock}). With Hutchinson's estimator on DiT, it spends $69\%$ in the Jacobian--vector product. The Broyden updates themselves are negligible when the transition is expensive. When it is cheap, as for the standard low-rank RNN, the low-rank scan dominates, and one Broyden iteration costs about $20\times$ a diagonal quasi-Newton iteration.

\begin{table}[H]
    \centering
    \caption{\textbf{Decomposition of the runtime ratio.} Runtime ratio $=$ iteration ratio $\times$ per-iteration cost ratio, baseline over Broyden variant ($>1$: the Broyden variant is faster). $T=10^5$ for the RNNs and the Gibbs sampler ($D=200$, with $\Delta t=3$ for the low-rank MLP-RNN and $\Delta t=0.5$ for the standard low-rank RNN). Iterations: baseline / method. For the RNNs, medians are over converged chains, so that iteration and runtime ratios refer to the same chains. Figure~\ref{fig:fig1} instead counts non-converged chains at the iteration cap. For the Gibbs sampler, medians are over all chains, and for DiT, means are over the $32$ images, as in the corresponding figures.}
    \label{tab:decomposition}
    \resizebox{\linewidth}{!}{%
    \begin{tabular}{lllrrrr}
        \toprule
        Benchmark & Baseline & Method & Iterations & Iter.\ ratio & Per-iter.\ ratio & Runtime ratio \\
        \midrule
        Low-rank MLP-RNN       & Diag.\ QN & PB & 38 / 19 & 2.00 & 15.74 & 31.48 \\
                               & Diag.\ QN & Accel.\ PB & 38 / 12 & 3.17 & 17.78 & 56.31 \\
                               & Newton & PB & 3 / 19 & 0.16 & 19.38 & 3.06 \\
                               & Newton & Accel.\ PB & 3 / 12 & 0.25 & 21.89 & 5.47 \\
        \midrule
        Standard low-rank RNN  & Diag.\ QN & PB & 30 / 14.5 & 2.07 & 0.04 & 0.09 \\
                               & Diag.\ QN & Accel.\ PB & 30 / 12 & 2.50 & 0.05 & 0.13 \\
                               & Newton & PB & 4 / 14.5 & 0.28 & 4.13 & 1.14 \\
                               & Newton & Accel.\ PB & 4 / 12 & 0.33 & 4.89 & 1.63 \\
        \midrule
        Gibbs, $S=100$         & Diag.\ QN & PB & 223.5 / 168 & 1.33 & 3.53 & 4.70 \\
                               & Diag.\ QN & Accel.\ PB & 223.5 / 40 & 5.59 & 5.32 & 29.70 \\
                               & Newton & PB & 9.5 / 168 & 0.06 & 15.29 & 0.86 \\
                               & Newton & Accel.\ PB & 9.5 / 40 & 0.24 & 23.02 & 5.47 \\
        \midrule
        Gibbs, $S=200$         & Diag.\ QN & PB & 164.5 / 118 & 1.39 & 5.46 & 7.61 \\
                               & Diag.\ QN & Accel.\ PB & 164.5 / 30.5 & 5.39 & 7.53 & 40.62 \\
        \midrule
        DiT                    & Diag.\ QN (Hutch.) & PB & 16.78 / 14.91 & 1.13 & 2.95 & 3.32 \\
                               & Diag.\ QN (Hutch.) & Accel.\ PB & 16.78 / 14.94 & 1.12 & 2.94 & 3.30 \\
        \bottomrule
    \end{tabular}}
\end{table}

\begin{table}[H]
    \centering
    \caption{\textbf{Wall-clock breakdown of one iteration} (H100, \texttt{float32}, $T=10^5$ except DiT). Measured time per iteration, and share of each component: evaluations of the transition $f$ (for DiT, the denoiser), automatic differentiation (Jacobian or its diagonal by reverse mode, or Hutchinson Jacobian--vector products for DiT), Broyden update (rank-one updates or refit of $U_t$), and parallel scan (with rank compression for the Broyden variants). Components are timed in isolation. Shares are fractions of their summed time, and the last column compares this sum with the measured iteration, in which XLA fuses part of the work (on the Gibbs sampler, the isolated components of PB and Accel.\ PB take the same time, but the measured Accel.\ PB iteration is faster, also in the sweep: $5.4$ vs.\ $8.2$\,ms). Matmul precision as in the corresponding experiments (DiT: TF32), so the measured times match the sweeps.}
    \label{tab:wallclock}
    \resizebox{\linewidth}{!}{%
    \begin{tabular}{llrrrrrr}
        \toprule
        Benchmark & Method & ms/iter. & $f$ (\%) & Autodiff (\%) & Update (\%) & Scan (\%) & Sum/meas. \\
        \midrule
        Low-rank MLP-RNN       & Diag.\ QN & 438 & 1 & 99 & -- & $<1$ & 1.00 \\
                               & Newton & 538 & $<1$ & 82 & -- & 17 & 1.00 \\
                               & PB & 27.9 & 22 & -- & 5 & 73 & 0.84 \\
                               & Accel.\ PB & 25.3 & 19 & -- & 17 & 64 & 1.06 \\
        \midrule
        Standard low-rank RNN  & Diag.\ QN & 1.05 & 51 & 14 & -- & 35 & 1.26 \\
                               & Newton & 90.9 & $<1$ & 6 & -- & 93 & 1.05 \\
                               & PB & 23.3 & 4 & -- & 6 & 90 & 0.82 \\
                               & Accel.\ PB & 19.8 & 3 & -- & 20 & 77 & 1.13 \\
        \midrule
        Gibbs, $S=100$         & Diag.\ QN & 29.4 & 8 & 91 & -- & 1 & 1.07 \\
                               & Newton & 127 & 2 & 26 & -- & 73 & 1.02 \\
                               & PB & 8.52 & 27 & -- & 9 & 64 & 1.04 \\
                               & Accel.\ PB & 5.66 & 27 & -- & 9 & 64 & 1.57 \\
        \midrule
        DiT                    & Diag.\ QN (Hutch.) & 418 & 31 & 69 & -- & $<1$ & 0.97 \\
                               & PB & 128 & 99 & -- & $<1$ & $<1$ & 1.00 \\
                               & Accel.\ PB & 128 & 99 & -- & $<1$ & $<1$ & 0.99 \\
        \bottomrule
    \end{tabular}}
\end{table}


\end{document}